\documentclass[lettersize,journal]{IEEEtran}
\usepackage{amsmath,amsfonts}
\usepackage{algorithmic}
\usepackage{algorithm}
\usepackage{array}
\usepackage[caption=false,font=normalsize,labelfont=sf,textfont=sf]{subfig}
\usepackage{textcomp}
\usepackage{stfloats}
\usepackage{url}
\usepackage{verbatim}
\usepackage{graphicx}
\usepackage{cite}
\usepackage{multicol}
\usepackage{multirow}
\usepackage{xspace} % 加载 xspace 宏包
\usepackage[dvipsnames,table,xcdraw]{xcolor} % 加载 xcolor 宏包并指定选项

\usepackage{amsmath}
\usepackage{booktabs}

\newcommand{\eg}{e.g.\xspace}
\newcommand{\ie}{i.e.\xspace}

\def\dq{\textcolor{black}}

\makeatletter
\def\ps@IEEEtitlepagestyle{
  \def\@oddfoot{\mycopyrightnotice}
  \def\@evenfoot{}
}
\def\mycopyrightnotice{
  {\footnotesize
  \begin{minipage}{\textwidth}
  \centering
    Copyright~\copyright~2025 IEEE. Personal use of this material is permitted. However, permission to use this material for any other purposes must be obtained from the IEEE by sending an email to pubs-permissions@ieee.org.
  \end{minipage}
  }
}

\begin{document}

\title{Exploring Audio Cues for Enhanced Test-Time Video Model Adaptation}
% \title{Using LLMs to Extract Audio Self-Supervisions for Enhanced Test-Time Video Model Adaptation}

\author{Runhao Zeng, Qi Deng, Ronghao Zhang, Shuaicheng Niu, Jian Chen, Xiping Hu, Victor C. M. Leung
        % <-this % stops a space
% \thanks{This work was supported by the National Natural Science Foundation of China (NSFC) under Grants 62202311; The Shenzhen Natural Science Foundation (the Stable Support Plan Program) under Grant 20220809180405001; Excellent Science and Technology Creative Talent Training Program of Shenzhen Municipality under Grant RCBS20221008093224017; The Guangdong Basic and Applied Basic Research Foundation under Grants 2023A1515011512.}
\thanks{This work was supported by the National Natural Science Foundation of China (NSFC) (Grant Nos. 62202311, 62376099), Excellent Science and Technology Creative Talent Training Program of Shenzhen Municipality (Grant  No. RCBS20221008093224017), Guangdong Basic and Applied Basic Research Foundation (Grant No. 2023A1515011512), Key Scientific Research Project of the Department of Education of Guangdong Province (Grant No. 2024ZDZX3012), and Natural Science Foundation of Guangdong Province (Grant No. 2024A1515010989). (\textit{Runhao Zeng and Qi Deng contributed equally to this work.}) (\textit{Corresponding Authors: Shuaicheng Niu, Jian Chen})}
\thanks{Runhao Zeng, Xiping Hu and Victor C. M. Leung are with the Artificial Intelligence Research Institute, Shenzhen MSU-BIT University and Guangdong-Hong Kong-Macao Joint Laboratory for Emotional Intelligence and Pervasive Computing, Shenzhen, 518172, China. Qi Deng, Ronghao Zhang and Jian Chen are with School of Software Engineering, South China University of Technology, Guangzhou, 510000, China. Shuaicheng Niu is with College of Computing and Data Science, Nanyang Technological University, 639798, Singapore. E-mail: runhaozeng.cs@gmail.com; dengqi.kei@gmail.com; zhangronghao16@gmail.com; shuaicheng.niu@ntu.edu.sg; ellachen@scut.edu.cn; huxp@bit.edu.cn; vleung@ieee.org}}

\markboth{IEEE Transactions on Circuits and Systems for Video Technology}%
{Shell \MakeLowercase{\textit{et al.}}: A Sample Article Using IEEEtran.cls for IEEE Journals}


\maketitle

\begin{abstract}
Test-time adaptation (TTA) aims to boost the generalization capability of a trained model by conducting self-/unsupervised learning during the testing phase. While most existing TTA methods for video primarily utilize visual supervisory signals, they often overlook the potential contribution of inherent audio data. To address this gap, we propose a novel approach that incorporates audio information into video TTA. Our method capitalizes on the rich semantic content of audio to generate audio-assisted pseudo-labels, a new concept in the context of video TTA. Specifically, we propose an audio-to-video label mapping method by first employing pre-trained audio models to classify audio signals extracted from videos and then mapping the audio-based predictions to video label spaces through large language models, thereby establishing a connection between the audio categories and video labels. To effectively leverage the generated pseudo-labels, we present a flexible adaptation cycle that determines the optimal number of adaptation iterations for each sample, based on changes in loss and consistency across different views. This enables a customized adaptation process for each sample. Experimental results on two widely used datasets (UCF101-C and Kinetics-Sounds-C), as well as on two newly constructed audio-video TTA datasets (AVE-C and AVMIT-C) with various corruption types, demonstrate the superiority of our approach. Our method consistently improves adaptation performance across different video classification models and represents a significant step forward in integrating audio information into video TTA. Code: https://github.com/keikeiqi/Audio-Assisted-TTA.
\end{abstract}


\section{Introduction}
\label{sec:intro}

\IEEEPARstart{D}{eep} neural networks have achieved significant success in various video analysis tasks~\cite{wang2021action,xiang2022spatiotemporal,yang2022recurring,zeng2021graph}, but most methods assume that training and testing data come from the same distribution. This assumption often fails in real-world scenarios, where distribution shifts occur due to environmental factors such as lighting, geographic location, and camera models. As a result, there has been growing interest in test-time adaptation (TTA) methods, which aim to adapt to distribution shifts by leveraging test samples.

\begin{figure}[!t]
    \centering 
    \includegraphics[width=\linewidth]{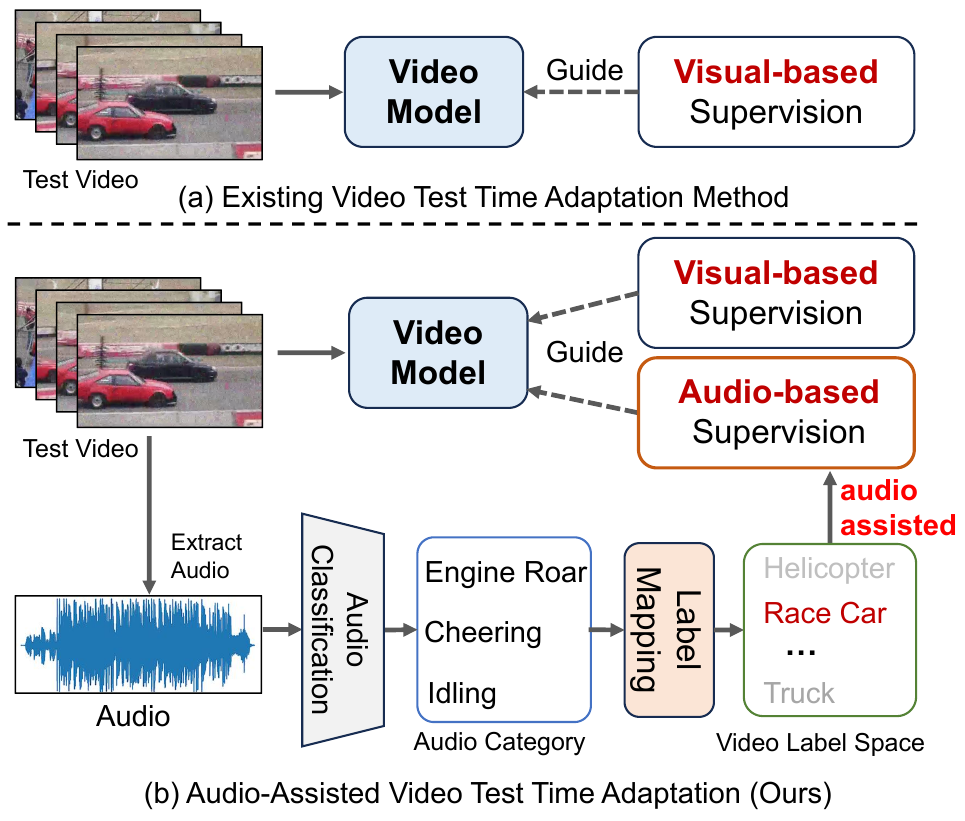}
     %\vspace{-0.6cm}
    \caption{Existing video test-time adaptation methods rely on visual supervision, overlooking the rich information inherent in audio. We propose a novel approach that involves extracting audio from videos and mapping the results of an open-source audio model to the video label space. This process generates audio-assisted pseudo labels, which significantly enhance TTA’s performance.}
    % \caption{Comparison with existing visual-based video TTA method.}
    \label{Fig1}
    %\vspace{-0.5cm}
\end{figure}

Although TTA methods have demonstrated effectiveness in image-based tasks~\cite{TENT,MEMO,T3A,DUA,LAME,gandelsman,niu2022efficient,TTT,TTT++}, their performance in video tasks remains limited~\cite{vitta}. A key challenge in video TTA is that video models process a sequence of frames, and when visual corruption occurs, it disrupts both the sequential information and temporal coherence essential for accurate predictions, leading to a significant degradation in performance~\cite{AME}. Recently, researchers have developed TTA methods tailored to video data~\cite{teco,AME,vitta}. As shown in Fig~\ref{Fig1}(a), these methods predominantly rely on visual information to obtain supervision (e.g., pseudo video label), neglecting the potential of audio as a supplementary modality.

\begin{figure}[!t]
    \centering 
    \includegraphics[width=\linewidth]{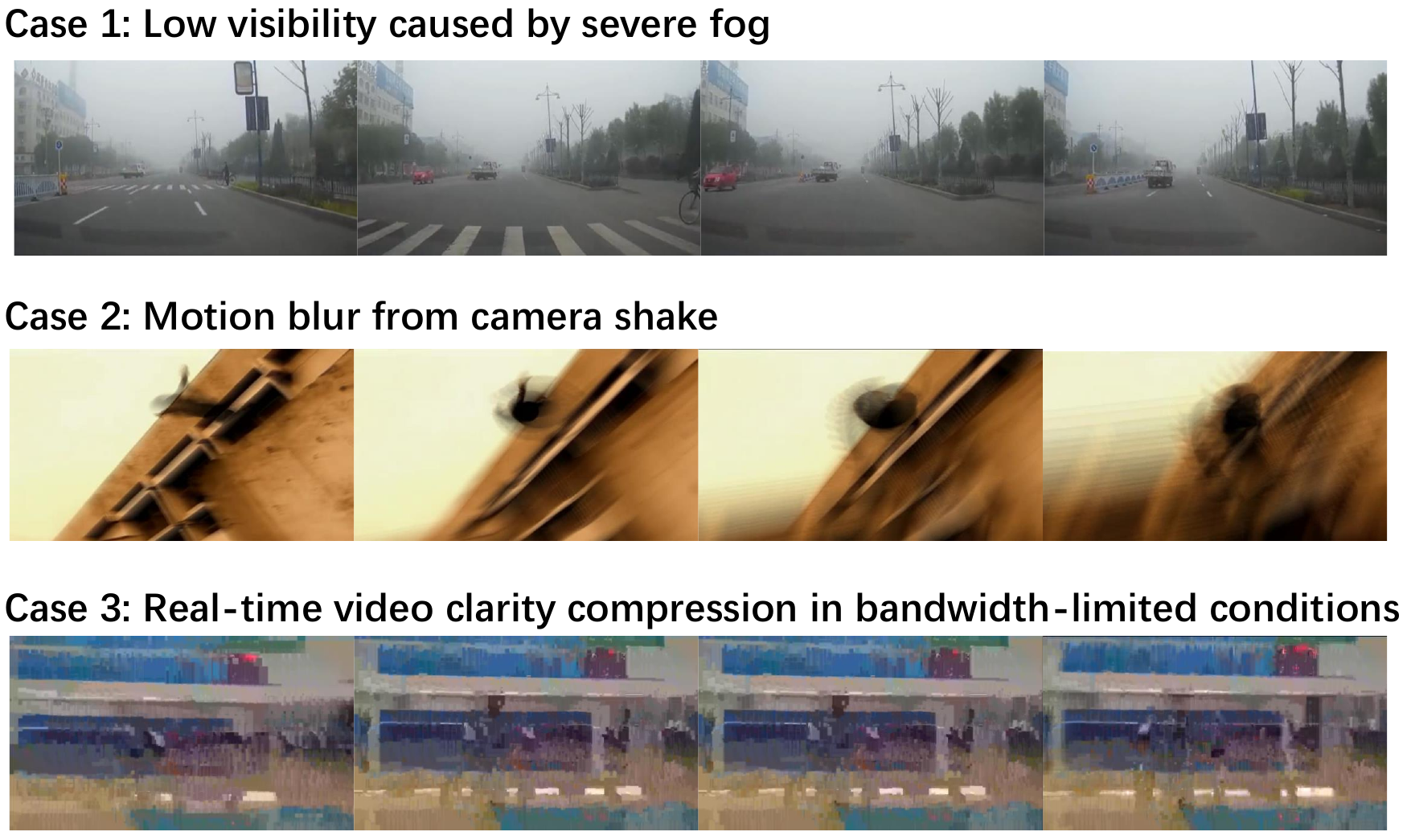}
    \caption{Common scenarios of video disruption include challenging environmental conditions (Case 1), camera motion (Case 2), and transmission issues (Case 3). In these situations, the primary impact is on the visual content, while the audio remains relatively unaffected. Since video models tend to perform poorly when visual quality is degraded, combining both video and audio modalities offers a more robust solution.}
    % \caption{Common scenarios of video disruption.}
    \label{fig_motivation_case}
    %\vspace{-0.5cm}
\end{figure}

In video data, audio and visual elements are often tightly synchronized and complementary. Human perception integrates both sensory modalities to better understand the environment. This becomes particularly relevant in real-world scenarios, where visual information may be degraded—due to adverse weather conditions, motion blur, or bandwidth limitations—while the audio stream remains relatively unaffected. For instance, as shown in Fig.~\ref{fig_motivation_case}, in conditions such as heavy fog or haze, visual sensors (e.g., cameras) may capture blurred or unrecognizable images, whereas the audio signal can still provide clear and informative content. Similarly, in fast-paced activities like parkour, motion blur can obscure visual details, yet the accompanying sounds (e.g., footsteps, environmental noises) remain informative. Additionally, in live video streaming, bandwidth constraints often result in compressed, low-quality video, while the audio stream may remain unaffected. These scenarios demonstrate that when visual data is compromised, audio can play a crucial role in enhancing the robustness of video classification models by providing complementary information.

However, integrating audio into TTA presents several challenges. In typical TTA settings, test data is unlabeled, making it impossible to directly train an audio classifier to predict video labels. A key challenge, therefore, is how to extract useful pseudo-labels from the audio to guide the adaptation of the video model. This issue is further complicated by the fact that audio and video label spaces are often mismatched, making it difficult to align audio-derived predictions with video action categories. For instance, as shown in Fig~\ref{Fig1}, when the audio contains predictions such as Engine Roar, Cheering and Idling, the corresponding video scene may involve actions like Race Car or Helicopter. To address this, a critical task is developing methods to map audio-derived pseudo-labels to the video label space without requiring additional training.

In this paper, we propose an audio-assisted approach for TTA in video models by leveraging the rich information embedded in audio, as illustrated in Fig \ref{Fig1}. Our approach consists of two main components: audio-video label mapping and a flexible adaptation cycle. \textbf{First}, we utilize existing pre-trained audio models, such as AST~\cite{gong2021ast}, to obtain audio categories. However, the predicted audio categories are not directly aligned with the desired video categories due to differences in the label spaces of the two modalities. Fortunately, we observe that both audio and video labels are semantically rich textual representations. Therefore, we propose a \textbf{audio-video label mapping} method based on large language models (LLMs), such as BERT~\cite{kenton2019bert} and GPT~\cite{brown2020language}. Specifically, we treat the predicted audio categories as text-related information relevant to the video and employ carefully designed prompting techniques. By leveraging the text comprehension abilities and knowledge inference capabilities of LLMs, we map these predicted audio categories to corresponding video labels. This enables the generation of pseudo video labels, which can serve as effective supervision for TTA.

\textbf{Second}, another challenge is the efficient utilization of the generated pseudo video labels. Our empirical findings indicate that performing TTA once per sample does not fully exploit the potential of audio-assisted pseudo labels. Performance can be further improved by repeating the adaptation for each sample. However, the optimal number of repetitions varies depending on the sample and the type of noise present. To address this, we propose a \textbf{flexible adaptation cycle} method, which designs adaptation conditions tailored to videos. This method uses the loss change between two consecutive adaptations and the consistency across different views of the same sample to determine the number of adaptations required. This allows for the adaptive selection of how many times pseudo video labels should be applied to each sample, ultimately achieving efficient audio-assisted TTA.

We construct two audio-video TTA datasets, AVE-C and AVMIT-C, by applying twelve common types of corruption to the videos. Additionally, we conduct experiments on video classification benchmarks, including UCF101-C and Kinetics-Sounds-C. Experimental results demonstrate that our method consistently improves adaptation performance across various backbone networks. Our main contributions are as follows:

\begin{itemize}
    \item To our best knowledge, we are the first to explore integrating inherent audio from videos to boost the performance of a trained video model on out-of-distribution test samples via TTA. With the help of long-neglected audios, our proposed realistic but challenging paradigm can bring significant improvements over existing video TTA methods, \eg, 27\% accuracy improvement over ViTTA on AVMIT-C dataset and TANet backbone.
    \item To effectively leverage audios, we propose a simple yet effective approach by utilizing Large Language Models to convert/align audio category space into any desired video label space to accommodate various datasets, without any additional training. To fully leverage audio-compensated labels, we devise a flexible adaptation cycle to dynamically determine the learning step of TTA for each test sample, further enhancing TTA performance. 
    \item We achieve significant improvements over the state-of-the-art video TTA methods on not only the existing action recognition benchmarks (\ie, Kinetics-Sounds-C and UCF101-C) but also two novel audio-video TTA datasets (\ie, AVMIT-C and AVE-C) we develop. The efficacy and generality of our method have been tested across these four datasets and various backbones.
   
\end{itemize}

\section{Related Work}
\label{sec:related_work}

\subsection{Test-Time Adaptation.}
Recent advances in video models, particularly in human-centric action recognition~\cite{guo2024benchmarking,li2025repetitive,li2024prototypical}, have achieved remarkable accuracy through specialized architectures and temporal modeling. However, these models often struggle with out-of-distribution (OOD) scenarios encountered in real-world settings, highlighting the need for effective TTA techniques.
Test-time adaptation (TTA) tackles the adaptation to unknown distribution shifts at test time in an unsupervised manner.
In recent years, it has garnered significant interest in the image domain~\cite{iwasawa2021test, niu2022efficient, su2022revisiting, nado2020evaluating, khurana2021sita, boudiaf2022parameter,gandelsman, chen2023camera, liu2024question, wu2024ttagaze}. One typical direction is to utilize a self-supervised auxiliary task to mitigate distribution shifts. For instance, TTT~\cite{TTT} adopted the self-supervised rotation prediction as an auxiliary task. TTT++~\cite{TTT++} proposed to use self-supervised contrastive learning. TENT~\cite{TENT} fine-tuned batch normalization’s affine parameters using an entropy penalty, whereas MEMO~\cite{MEMO} adapted the network at test-time by minimizing the entropy of the marginal output distribution across augmentations. Although existing TTA methods achieve good results in image-based tasks, they do not perform as well in video-based tasks. Recently, TeCo~\cite{teco} minimized the entropy of the prediction based on the global video content and fed local content to regularize the temporal coherence. AME~\cite{AME} proposed to exploit motion cues to enhance the motion encoding capability of video models. ViTTA~\cite{vitta} aligned the testing statistics of certain layers with those derived from the training set, achieving the current best performance in video TTA.
However, these methods largely depend on supervision derived from visual information, overlooking the rich data contained in audio.
In this paper, we present a viable approach for utilizing audio in video TTA and empirically demonstrate that considering audio can significantly enhance the performance of TTA.

\subsection{Audio-Visual Learning for Video Classification.}
Audio has been demonstrated to enhance action recognition performance beyond merely using visual modality alone~\cite{wang2020makes, alwassel2020self, arandjelovic2017look}. One of the main components of the audio-visual learning algorithm is modality fusion, which can be categorized as early, mid, and late fusion~\cite{kazakos2019epic, wang2016exploring, feichtenhofer2019slowfast}. 
Differing from these simple fusion strategies, the self-attention~\cite{vaswani2017attention} operation of transformers provides a natural mechanism to aggregate audio-visual information~\cite{likhosherstov2021polyvit, nagrani2021attention, chen2022mm}.
Meanwhile, recent video understanding works like TAMT~\cite{TAMT} demonstrated the effectiveness of multimodal fusion through timestamp-aligned visual frames and ASR transcripts for improved video summarization.
% 针对模态融合中存在的modality imbalance的问题,\cite{xx}进行了分析与研究。
\cite{chen2024bootstrapping, fu2023multimodal} conducted an analysis and research regarding the issue of modality imbalance in modal fusion.
Recently, some works~\cite{planamente2021cross, planamente2022domain, yang2021epic} considered that audio has less variance across domains to alleviate the domain shift in action recognition across scenes. 
Among the studies, the work most relevant to ours is~\cite{zhang2022audio}. They introduced an audio-infused recognizer to leverage domain-invariant cues inherent in the audio. However, their approach necessitates the pre-training of a model capable of taking audio input and producing corresponding video labels. This method is not feasible in our TTA setting, as we lack the audio-label pairs for training. In our paper, we study a scenario more aligned with practical applications. Given any pre-trained video model, we propose a technique for extracting pseudo video labels from audio to facilitate TTA.

\section{Proposed Method}
\label{sec:proposed_method}

\subsection{Problem Statement} Consider $P(v)$ as the distribution of a set of source training videos $\{v_n\}_{n=1}^{N}$, where each video $v_n$ follows the distribution $P(v)$. A model, denoted as $f_\Theta(\cdot)$ and parameterized by \( \Theta \), is trained using labeled video samples $\{v_n, y_n\}_{n=1}^{N}$. The corresponding label for each video is $y_n \in \mathcal{Y}_v$, representing the label space. In an ideal scenario, the model $f_\Theta(\cdot)$ should perform effectively on test samples that are drawn from the same distribution as the training set, namely $v\sim P(v)$. However, test samples often exhibit variations or corruptions not present in the training set, leading to out-of-distribution samples, denoted as $v\sim U(v)$, where $U(v)\neq P(v)$. This discrepancy typically results in substantial performance degradation. 

To enhance the model's generalization to out-of-distribution test samples, video test-time adaptation (TTA)~\cite{vitta} involves unsupervised learning at test time, utilizing only unlabeled test samples $v$ to update the model $f_\Theta(\cdot)$ in an online manner.

% To enhance the model's generalization to out-of-distribution test samples, the technique of video test-time adaptation (ViTTA), as introduced in~\cite{vitta}, has gained prominence. ViTTA involves unsupervised learning at test time, utilizing only unlabeled test samples $v$ in an online manner. However, such a method primarily considers visual information and overlooks the rich audio information in videos. This type of information, less susceptible to visual noise, often provides valuable cues, such as object and scene information, aiding in the extraction of effective supervisory under testing scenarios.

\subsection{General Scheme}
We address the problem that existing video TTA methods primarily focus on visual information while neglecting the valuable audio cues present in videos. Audio, being less prone to visual noise, often contains important contextual information, such as object and scene cues, which can enhance the supervisory signals during testing. In this study, we propose leveraging the inherent audio signals in videos to improve the performance of video TTA.

\begin{figure*}[!t]
    \centering  
    \includegraphics[width=\textwidth]{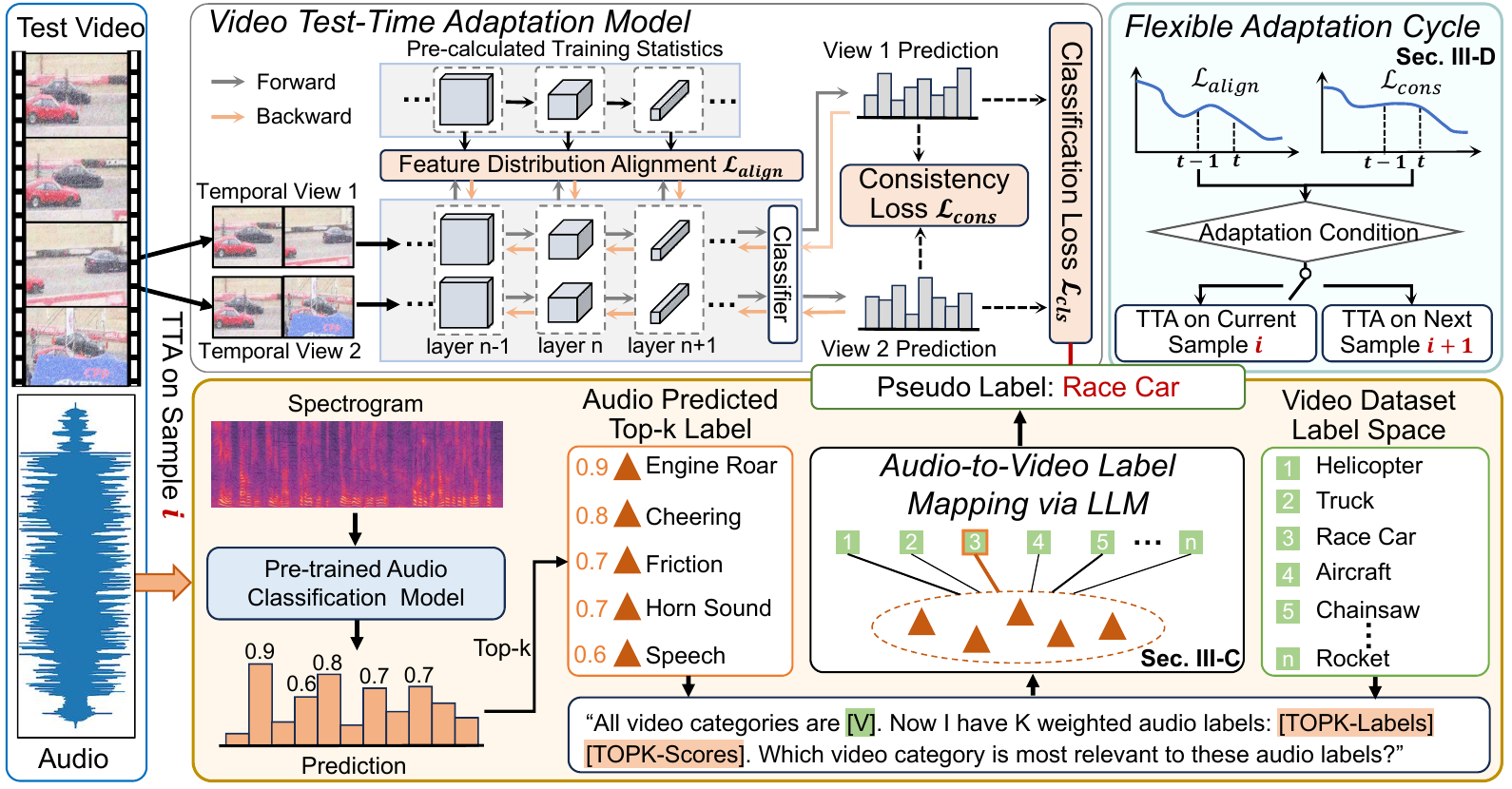} 
    \caption{Overall illustration of our proposed method. Given a paired video-audio test sample, we first extract audio labels using a pre-trained audio classification model and then align/map these audio labels with/to the video label space through a devised large language model (LLM) prompting technique. Next, we calculate the classification loss using these audio-assisted pseudo video labels for video TTA. Additionally, we propose a flexible adaptation cycle scheme to further boost adaptation performance by actively adjusting the learning iteration steps for each test sample based on a devised adaptation condition.}
    % \caption{Overall illustration of our proposed method.}
    \label{Fig.2} 
    %\vspace{-0.4cm}
\end{figure*}

Without loss of generality, given a video $v$, we forward it to a pre-trained video model to obtain classification prediction $\hat{y}_v$. To conduct TTA, we need a pseudo label $\bar{y}$ and optimize the following loss function:
\begin{eqnarray}
            \mathcal{L}_{cls}=-\sum\limits_{c=1}^C \bar{y}^c\log \hat{y}_v^c.
            \label{audio_video_ce}
\end{eqnarray}
% $\mathbb{I}$ is the indicator function, being 1 if xxxx and 0 otherwise.
Unlike existing methods that derive \(\bar{y}\) from visual information, we propose using audio $a$ extracted from videos to acquire reliable pseudo labels. To this end, we propose to exploit an open-source pre-trained audio classification model $f_A$, \eg, AST~\cite{gong2021ast} to predict audio category $\bar{y}_a$, which are typically sound-related (\eg, crowd, water, fire, bus), differing completely from the video label space. Thus, $\bar{y}_a$ cannot be directly used in Eqn.(\ref{audio_video_ce}). To address this, we devise a \textbf{LLM-based Audio-to-Video Label Mapping} method to predict pseudo video label $\bar{y}_v$ relying on the semantic relations between audio and video categories via
\begin{eqnarray}
            \bar{y}_v=f_L(P(f_A(a),\mathcal{Y}_v)),
\end{eqnarray}
where $P$ is a prompt generator, $\mathcal{Y}_v$ is the video label space, and $f_L$ can be any large language model such as BERT~\cite{kenton2019bert}, GPT~\cite{brown2020language}. Another challenge is how to use $\bar{y}_v$ for TTA. Since the number of adaptations varies per sample, 
% Existing methods adapt a single sample once, but our preliminary experiments suggest that some samples, when adapted multiple times, enhance model performance, and the number of adaptations varies per sample. 
we propose an \textbf{Adaptive Audio-Visual Adaptation} method. In particular, we assess the consistency of predictions between the two views (w.r.t the same sample) and observe the change in loss between two adaptation steps, to select the number of adaptation steps for each sample. 
After each adaptation of an input test sample, we input the visual data of this sample into the updated video model for prediction. \textit{In other words, the model, after undergoing adaptation through our method, can be deployed and make predictions under its original conditions (for example, using only visual data as input).} The schematic of our approach is shown in Fig~\ref{Fig.2}. We organize the sections as follows:

%每次对一个input test sample 进行adaptation后，我们将该sample的visual data输入到update后的video model进行预测。也就是说，经过我们方法adapt之后的模型可以在它原来的应用条件下（例如，只将visual data作为输入）进行部署和预测

\begin{itemize}

\item Sec.~\ref{Sec:LLM} details how to employ LLM to map audio labels to video labels, addressing the issue of label space mismatch between the audio and video labels.

\item Sec.~\ref{Sec:Adaptive} introduces an additional model adaptation strategy that adaptively determines the number of adaptation steps for each sample, thereby enhancing the utilization of pseudo-video labels.

\end{itemize}

% In the following, we aim to answer two questions: (1) how to obtain pseudo video label relying on audio under TTA setting, and (2) how to adaptively use the audio-assisted video label for each sample.

\subsection{Audio-to-Video Label Mapping via LLMs}\label{Sec:LLM}
The key challenge to obtain pseudo video label from audio input lies in the unknown labels of video and audio, compounded by the absence of an audio classifier corresponding to the video label space for direct prediction based on audio content. Training a dataset-specific audio classification model with a label space identical to the video label space is resource-consuming and, more importantly, infeasible under the TTA setting. 

To classify audio, we can leverage open-source pre-trained audio classification models $f_A$ (\eg, AST~\cite{gong2021ast}). However, the audio category obtained through this method cannot directly provide the corresponding video category. Therefore, we propose a pseudo label mapping method based on large language models (LLMs), utilizing LLMs $f_L$ to infer the latent semantic association between audio categories and video labels, thereby deriving pseudo video labels. The key to exploiting LLMs lies in designing an efficient and reasonable prompt. Instead of simply inputting the audio prediction results and the labels contained in the corresponding video label space into the LLMs, we propose a template-based prompt generation module (PGM) $f_P$. Specifically, we select the top five categories with the highest scores in the predictions from $f_A$. Since the probability assigned to each category also reflects the confidence level, this is a crucial piece of information. We design a text prompt with five modules: Background, Task, Examples, Requirements, and Inputs, as follows:
\textit{\#Human: ``\textless Background Introduction\textgreater  \textless Task Description\textgreater \textless Requirements\textgreater \textless Examples\textgreater  \textless \textbf{Inputs}: video label space [V], audio predictions [TOPK-Labels] [TOPK-Scores]\textgreater .'' \#Assistant:},
where [TOPK-Labels] and [TOPK-Scores] denote the top-K predictions of the audio classification model and their predicted probability or confidence corresponding to each audio label, respectively. For a comprehensive view of the prompt, please refer to Fig. \ref{Alg:prompt}.

Overall, we input the video label space, the top five predictions of the audio model, and their probabilities into PGM and compute
\begin{eqnarray}
            \bar{y}_v=f_L(f_P(\mathcal{Y}_v, \bar{y}_a, \hat{p}_a)).
            \label{LLM_predict}
\end{eqnarray}
% We empirically found that LLMs may sometimes fail to find the video label. In such cases, we opt to repeatedly utilize the LLMs, selecting the label that appears most frequently as the final corresponding pseudo video label. 
We sample frames from the same video for $M$ times and forward them to the video model, lastly adapt the model by optimizing the following loss function
\begin{eqnarray}
            \mathcal{L}_{cls}=-\sum\limits_{m=1}^M \bar{y}_v\log \hat{y}_v^m.
            \label{cls}
\end{eqnarray}
Note that we use only existing open-source models, obviating the need for additional training, which not only saves training costs but also aligns more closely with real-world application scenarios. Furthermore, this method has broad applicability and can be transferred to different datasets.

\subsection{Adaptive Audio-Visual Adaptation}\label{Sec:Adaptive}
With the audio-assisted pseudo video label, one can utilize the loss calculated by Eqn. (\ref{cls}) to update the model. Then, proceed to the forward and backward adaptation of the next sample, a process we name single-step adaptation.

\subsubsection{Single-Step Audio-Visual Adaptation} 
Besides acquiring supervision through audio as previously mentioned, we aim to further leverage supervision from the visual content. Specifically, we consider two types of loss, including cross-view consistency loss to exploit the relations between different views of the same sample and feature alignment loss to leverage the knowledge in the training set.

\noindent \textbf{Cross-view consistency loss.} We enforce consistency among the corresponding predictions of the $M$ views. The learning target is obtained by averaging the class probabilities predicted by the video model for the input views, and the cross-view consistency loss is defined as
\begin{eqnarray}
            \mathcal{L}_{cons}=\sum\limits_{m=1}^M\left|\hat{y}_v^m-\frac{1}{M} \sum\limits_{j=1}^M \hat{y}_v^j\right|.
            \label{consistency}
\end{eqnarray}

\noindent \textbf{Feature alignment loss.} We denote the feature maps from the $l$-th layer as $g_l(v)\in \mathbb{R}^{c_l \times t_l \times h_l \times w_l}$ respectively, $c_l$ is the number of channels in the $l$-th layer, $t_l$ refers to the temporal dimension, and $h_l$ and $w_l$ represents the spatial dimensions. Hence, we compute the mean vectors of the $l$-th layer features by
\begin{equation}
% \begin{aligned}
    \mu_l(v) = \mathbb{E}_{(i,j,z) \in V}[g_l(v)[:,i,j,z]], \label{con:mean}
% \end{aligned}  
\end{equation}
where $V = [1,t_l]\times[1,h_l]\times[1,w_l]$.
Based on the mean vectors, the variance vectors of the $l$-th layer features can be obtained by calculating
\begin{equation}
\begin{aligned}
    \sigma_l^{2}(v) = \mathbb{E}_{(i}& _{,j,z) \in V}[(g_l(v)[:,i,j,z] - \mu_{l}(v))^2].
    \label{con:var}
\end{aligned}  
\end{equation}
Our target is to align the testing statistics of selected layers to the statistics pre-calculated from the training data, by minimizing the following objective
\begin{eqnarray}
            \mathcal{L}_{align}=\sum\limits_{l\in L}\left|\mu_l(T)-\hat{\mu}_l\right|+\left| \sigma^2_l(T)-\hat{\sigma}^2_l\right|,
            \label{alignment}
\end{eqnarray}
where $L$ is the set of layers to be aligned, $\left| \cdot\right|$ denotes the vector $l_1$ norm, recall that $T$ denotes the test set, $\hat{\mu}_l$ and $\hat{\sigma}^2_l$ are pre-calculated training statistics. More details about $\mathcal{L}_{cons}$ and $\mathcal{L}_{align}$ please refer to~\cite{vitta}. The overall loss function of our audio-visual adaptation approach is
\begin{eqnarray}
\mathcal{L}=\alpha\mathcal{L}_{cls}+\beta\mathcal{L}_{cons}+\mathcal{L}_{align},
\label{Eq:total_loss}
\end{eqnarray}
where $\alpha$ and $\beta$ are hyperparameters to trade off these losses. We set $\alpha=\beta=0.1$ in all the experiments and find that it works well across all of them.
% where $\alpha$ and $\beta$ are hyper-parameters to trade off these losses. We simply set $\beta$=0.1(following ViTTA) and $\alpha$=0.1 (to maintain the same loss magnitude to $\mathcal{L}_{cons}$) in all the experiments and find that it works well across all of them.

\subsubsection{Multi-Step Adaptation via Flexible Adaptation Cycle}
We empirically found that adapting each sample only once may not yield good results (see Fig~\ref{Fig4}). Considering minor, even negligible, increases in time, significant gains can be achieved by repetitively adapting the same sample, thereby attaining better TTA results. However, simply allowing all samples to be adapted multiple times is impractical.
We found that the number of adaptations required for each sample is dependent on the sample itself and the type of corruption it has undergone. While some samples may achieve excellent adaptation with very few rounds, others may necessitate a greater number. To address this, we propose a method named flexible adaptation cycle, which automatically selects the number of adaptation steps for each sample.
% \noindent \textbf{Flexible Adaptation Cycle.}

\begin{algorithm}[!t]
    \caption{Our Audio-Assisted Video TTA Method}
    % \begin{flushleft}
    % \end{flushleft}
    \begin{algorithmic}[1]
    \REQUIRE{Test samples $\{v_n\}_{n=1}^{N_t}$, video label space $\mathcal{Y}_v$, pre-trained video model $f_\Theta$ with parameters $\Theta$.}
        \FOR {$i=1\dots N_t$}
            \STATE // \textbf{\textit{Get audio prediction}}
            \STATE Extract audio $a_i$ from video $v_i$.
            \STATE Obtain top-K audio labels $\bar{y}_a$ and probability $\hat{p}_a$ using an open-source pre-trained audio model.
            \STATE // \textbf{\textit{Pseudo Label Mapping via LLMs}}
            \STATE Get pseudo video label $\bar{y}_v$ using LLMs via Eqn. (\ref{LLM_predict}).
            \STATE // \textit{Adaptive Audio-Visual Adaptation}
            \STATE Initialize repetition step $t=0$.
            \WHILE {$t<\tau$}
                \STATE Calculate classification loss $\mathcal{L}_{cls}$ relying on $\bar{y}_v$ via Eqn. (\ref{cls}).
                \STATE Calculate consistency loss $\mathcal{L}_{cons}$ via Eqn. (\ref{consistency}).
                \STATE Compute mean and variance of the test data feature map layers via Eqn. (\ref{con:mean}) and (\ref{con:var}).
                \STATE Calculate alignment loss $\mathcal{L}_{align}$ via Eqn. (\ref{alignment}).
                \STATE Update $\Theta$ by minimizing the loss in Eqn. (\ref{Eq:total_loss}).
                \STATE // \textit{Adaptation condition}
                \STATE Obtain $R^t$ to determine whether to continue to adapt after the $t$-th repetition steps via Eqn. (\ref{Eq:condition}).
                
                \STATE Let $t=t+1$ \textbf{if} $R^t=1$ \textbf{else} break
            \ENDWHILE
            \STATE // \textbf{\textit{Inference with visual data only}}
            \STATE Obtain video classification result: $\hat{y}_i = f_\Theta(v_i)$.
        \ENDFOR
    \end{algorithmic}
    % \begin{flushleft}
    %  \textbf{Output:} The predictions via Eqn.
    % \end{flushleft}
    \label{Alg:forward}
 \end{algorithm}

Our criteria for determining the adaptation cycle are based on the adaptation goal in Eqn. (\ref{Eq:total_loss}), with two primary considerations: the presence of a downward trend in loss and the consistency of predictions across different views. 
Our underlying rationale, taking $\mathcal{L}_{cons}$ as an illustration, is that for a given sample, if adaptation is executed consecutively twice and meets the criteria of $\mathcal{L}_{cons}^{(t)}<\mathcal{L}_{cons}^{(t-1)}$, this suggests a potential for further optimization, warranting continued adaptation using this sample. Regarding the two views of the same sample, if their predicted labels diverge $\hat{y}_i^{(t)} \neq \hat{y}_j^{(t)}$, we advocate for a new cycle of adaptation to this sample.
To ensure efficiency, we also set a hyperparameter $\tau$ that limits the maximum number of cycles. The criteria is defined as:
\begin{equation}
    R^t=\mathbb{I}_{\{\mathcal{L}_{cons}^{(t)}<\mathcal{L}_{cons}^{(t-1)} ~\textbf{or}~ \mathcal{L}_{align}^{(t)}<\mathcal{L}_{align}^{(t-1)}~\textbf{or}~ \hat{y}_i^{(t)} \neq \hat{y}_j^{(t)}\}} ,
    \label{Eq:condition}
\end{equation}
where $\mathbb{I}_{\{\cdot\}}$ is an indicator function and $R^t$ means whether to continue to adapt after the $t$-th cycle, $i,j\in\{1,2,...,M\}$. In brief, if the current sample satisfies the conditions in Eqn. (\ref{Eq:condition}), it should undergo one more adaptation cycle. Otherwise, the adaptation should advance to the next sample. The algorithmic depiction of our adaptation method is shown in Algorithm~\ref{Alg:forward}.

\begin{table*}[!t]
\caption{Comparison with state-of-the-art TTA methods on AVMIT-C and AVE-C dataset w.r.t. accuracy (\%).}
\centering
\resizebox{\textwidth}{!}{
    \begin{tabular}{lllcccccccccccc|c}
    \toprule
    \multirow{2}{*}{Dataset}  & \multirow{2}{*}{Backbone} & \multirow{2}{*}{Methods} & \multicolumn{13}{c}{Corruptions}                                                                                                                                                                                           \\  \cline{4-16} 
                              &                           &                          & Gauss          & Pepper         & Salt           & Shot           & Contrast       & Impulse        & Rain           & Zoom           & Motion         & Jpeg           & Defocus        & H265.abr       & Avg.           \\ \hline
    \multirow{18}{*}{AVMIT-C} & \multirow{9}{*}{TANet}    & Source                   & 41.46          & 36.77          & 32.19          & 71.04          & 17.19          & 41.98          & 50.21          & 45.63          & 76.15          & 75.21          & 61.46          & 63.54          & 51.07          \\
                              &                           & BN-Adapt~\cite{bn-adapt}                 & 41.35          & 38.13          & 35.94          & 69.17          & 25.94          & 41.35          & 54.27          & 39.27          & 70.94          & 71.67          & 58.13          & 57.29          & 50.29          \\
                              &                           & TENT~\cite{TENT}                     & 41.35          & 40.63          & 38.44          & 72.71          & 22.29          & 42.50          & 53.75          & 43.02          & 76.15          & 75.31          & 61.46          & 54.48          & 51.84          \\
                              &                           & SHOT~\cite{SHOT}                     & 43.44          & 41.25          & 39.69          & 71.56          & 26.98          & 44.90          & 58.54          & 44.17          & 75.94          & 74.06          & 62.08          & 62.29          & 53.74          \\
                              &                           & NORM~\cite{NORM}                     & 49.79          & 42.50          & 41.15          & 72.40          & 29.27          & 48.85          & 74.79          & 55.62          & 72.29          & 71.56          & 56.56          & 50.42          & 55.43          \\
                          &                           & \dq{DeYO~\cite{deyo}}                     & \dq{45.03}          & \dq{42.47}          & \dq{40.17}          & \dq{75.31}          & \dq{25.80}          & \dq{46.08}          & \dq{69.58}          & \dq{44.69}          & \dq{76.91}          & \dq{78.13}          & \dq{63.85}          & \dq{61.94}          & \dq{55.83}          \\
                          &                           & \dq{ROID~\cite{roid}}                     & \dq{50.63}          & \dq{44.48}          & \dq{36.67}          & \dq{60.31}          & \dq{19.79}          & \dq{51.67}          & \dq{59.38}          & \dq{55.52}          & \dq{81.15}          & \dq{64.58}          & \dq{75.52}          & \dq{69.48}          & \dq{55.76}          \\                              &                           & ViTTA\dq{~\cite{vitta}}                    & 52.71          & 49.58          & 38.65          & 77.19          & 37.08          & 52.71          & 70.21          & 55.63          & 76.56          & 77.08          & 65.42          & 64.27          & 59.76          \\
                              &                           & \cellcolor{gray!10}Ours                     & \cellcolor{gray!10}\textbf{72.40} & \cellcolor{gray!10}\textbf{63.85} & \cellcolor{gray!10}\textbf{65.73} & \cellcolor{gray!10}\textbf{84.06} & \cellcolor{gray!10}\textbf{63.65} & \cellcolor{gray!10}\textbf{71.88} & \cellcolor{gray!10}\textbf{80.94} & \cellcolor{gray!10}\textbf{72.81} & \cellcolor{gray!10}\textbf{83.75} & \cellcolor{gray!10}\textbf{86.56} & \cellcolor{gray!10}\textbf{78.85} & \cellcolor{gray!10}\textbf{72.92} & \cellcolor{gray!10}\textbf{74.78} \\ \cline{2-16} 
                              & \multirow{9}{*}{TSM}      & Source                   & 33.54          & 32.81          & 27.08          & 66.46          & 18.75          & 34.90          & 50.63          & 59.90          & 84.38          & 76.46          & 61.88          & 63.13          & 50.83          \\
                              &                           & BN-Adapt\dq{~\cite{bn-adapt}}                 & 36.67          & 37.19          & 33.44          & 67.40          & 30.73          & 38.75          & 62.50          & 56.46          & 80.94          & 72.60          & 60.52          & 57.71          & 52.91          \\
                              &                           & TENT\dq{~\cite{TENT}}                     & 39.38          & 37.71          & 37.92          & 72.08          & 24.27          & 41.67          & 59.79          & 58.85          & 84.69          & 76.67          & 65.63          & 60.63          & 54.94          \\
                              &                           & SHOT\dq{~\cite{SHOT}}                     & 40.83          & 38.23          & 39.37          & 68.75          & 35.52          & 42.81          & 65.21          & 58.23          & 82.40          & 74.17          & 63.44          & 61.46          & 55.87          \\
                              &                           & NORM\dq{~\cite{NORM}}                     & 40.00          & 42.50          & 34.12          & 64.79          & 25.63          & 49.27          & 67.40          & 49.27          & 64.48          & 64.17          & 55.73          & 45.52          & 50.24          \\
                          &                           & \dq{DeYO~\cite{deyo}}                     & \dq{36.15}          & \dq{36.77}          & \dq{33.93}          & \dq{69.38}          & \dq{29.79}          & \dq{38.65}          & \dq{68.41}          & \dq{60.89}          & \dq{84.53}          & \dq{76.51}          & \dq{65.26}          & \dq{63.07}          & \dq{55.28}          \\
                          &                           & \dq{ROID~\cite{roid}}                     & \dq{41.04}          & \dq{41.67}          & \dq{32.08}          & \dq{67.76}          & \dq{19.17}          & \dq{44.48}          & \dq{63.43}          & \dq{61.36}          & \dq{83.44}          & \dq{78.13}          & \dq{64.30}          & \dq{65.48}          & \dq{55.20}          \\                              &                           & ViTTA\dq{~\cite{vitta}}                    & 45.73          & 42.40          & 33.13          & 71.77          & 25.42          & 46.15          & 59.90          & 60.00          & 84.48          & 76.77          & 65.63          & 63.33          & 56.23          \\
                              &                           & \cellcolor{gray!10}Ours                     & \cellcolor{gray!10}\textbf{52.60} & \cellcolor{gray!10}\textbf{45.83} & \cellcolor{gray!10}\textbf{39.38} & \cellcolor{gray!10}\textbf{75.31} & \cellcolor{gray!10}\textbf{43.65} & \cellcolor{gray!10}\textbf{56.77} & \cellcolor{gray!10}\textbf{71.15} & \cellcolor{gray!10}\textbf{62.60} & \cellcolor{gray!10}\textbf{85.31} & \cellcolor{gray!10}\textbf{78.54} & \cellcolor{gray!10}\textbf{67.81} & \cellcolor{gray!10}\textbf{66.98} & \cellcolor{gray!10}\textbf{62.16} \\ \hline
    \multirow{18}{*}{AVE-C}   & \multirow{9}{*}{TANet}    & Source                   & 35.82          & 35.32          & 23.38          & 60.20          & 20.40          & 35.57          & 56.97          & 42.29          & 68.41          & 48.26          & 50.75          & 51.24          & 44.05          \\
                              &                           & BN-Adapt\dq{~\cite{bn-adapt}}                 & 27.36          & 32.59          & 21.39          & 59.20          & 13.68          & 27.36          & 61.69          & 42.79          & 65.42          & 47.26          & 49.75          & 44.53          & 41.09          \\
                              &                           & TENT\dq{~\cite{TENT}}                     & 24.13          & 30.85          & 18.41          & 60.70          & 19.40          & 22.89          & 61.44          & 43.28          & 67.66          & 48.76          & 55.47          & 46.52          & 41.63          \\
                              &                           & SHOT\dq{~\cite{SHOT}}                     & 42.79          & 43.28          & 34.08          & 62.44          & 28.36          & 46.02          & 62.69          & 49.00          & 64.68          & 52.49          & 47.51          & 50.75          & 48.67          \\
                              &                           & NORM\dq{~\cite{NORM}}                     & 38.06          & 38.81          & 31.34          & 59.95          & 20.90          & 40.55          & 64.18          & 50.25          & 63.68          & 59.45          & 45.52          & 42.54          & 46.27          \\
                          &                           & \dq{DeYO~\cite{deyo}}                     & \dq{39.55}          & \dq{35.57}          & \dq{25.87}          & \dq{64.93}          & \dq{20.65}          & \dq{41.29}          & \dq{64.93}          & \dq{43.53}          & \dq{67.91}          & \dq{53.23}          & \dq{53.48}          & \dq{51.00}          & \dq{46.83}          \\
                          &                           & \dq{ROID~\cite{roid}}                     & \dq{40.42}          & \dq{38.93}          & \dq{24.88}          & \dq{72.14}          & \dq{24.38}          & \dq{40.30}          & \dq{66.79}          & \dq{49.75}          & \dq{68.78}          & \dq{53.98}          & \dq{57.59}          & \dq{57.71}          & \dq{49.64}          \\
                              &                           & ViTTA\dq{~\cite{vitta}}                    & 51.00          & 53.23          & 36.07          & 69.65          & 36.07          & 51.54          & 64.43          & 48.01          & 69.65          & 67.66          & 51.24          & 53.73          & 54.36          \\
                              &                           & \cellcolor{gray!10}Ours                     & \cellcolor{gray!10}\textbf{62.44} & \cellcolor{gray!10}\textbf{63.68} & \cellcolor{gray!10}\textbf{51.74} & \cellcolor{gray!10}\textbf{74.63} & \cellcolor{gray!10}\textbf{47.51} & \cellcolor{gray!10}\textbf{59.70} & \cellcolor{gray!10}\textbf{70.65} & \cellcolor{gray!10}\textbf{62.44} & \cellcolor{gray!10}\textbf{70.40} & \cellcolor{gray!10}\textbf{70.15} & \cellcolor{gray!10}\textbf{61.94} & \cellcolor{gray!10}\textbf{58.21} & \cellcolor{gray!10}\textbf{62.79} \\ \cline{2-16} 
                              & \multirow{9}{*}{TSM}      & Source                   & 26.37          & 31.84          & 19.15          & 59.20          & 12.69          & 26.12          & 51.24          & 44.28          & 65.67          & 48.26          & 48.26          & 50.25          & 40.28          \\
                              &                           & BN-Adapt\dq{~\cite{bn-adapt}}                 & 29.10          & 34.33          & 24.38          & 56.22          & 17.41          & 29.60          & 60.20          & 47.26          & 67.16          & 47.02          & 53.23          & 46.77          & 42.72          \\
                              &                           & TENT\dq{~\cite{TENT}}                     & 27.86          & 32.34          & 23.13          & 59.45          & 19.40          & 27.11          & 60.45          & 49.01          & 66.67          & 48.76          & 55.22          & 46.27          & 42.97          \\
                              &                           & SHOT\dq{~\cite{SHOT}}                     & 39.30          & 39.05          & 28.86          & 62.94          & 21.14          & 31.34          & 44.28          & 42.04          & 40.55          & 47.51          & 38.31          & 50.75          & 40.51          \\
                              &                           & NORM\dq{~\cite{NORM}}                     & 37.81          & \textbf{48.26} & 24.88          & 64.18          & 7.71           & 35.57          & 61.44          & 47.51          & 60.70          & 64.43          & 47.01          & 48.01          & 45.63          \\
                          &                           & \dq{DeYO~\cite{deyo}}                     & \dq{30.22}          & \dq{34.45}          & \dq{24.73}          & \dq{59.58}          & \dq{23.22}          & \dq{32.09}          & \dq{60.35}          & \dq{49.50}          & \dq{66.24}          & \dq{51.74}          & \dq{52.92}          & \dq{50.75}          & \dq{44.65}          \\
                          &                           & \dq{ROID~\cite{roid}}                     & \dq{29.10}          & \dq{43.53}          & \dq{20.90}          & \dq{70.40}          & \dq{11.69}          & \dq{30.60}          & \dq{59.20}          & \dq{52.74}          & \dq{66.92}          & \dq{62.94}          & \dq{49.00}          & \dq{57.96}          & \dq{44.65}          \\
                              &                           & ViTTA\dq{~\cite{vitta}}                    & 41.54          & 43.04          & 26.87          & 65.17          & 23.88          & 38.81          & 64.93          & 45.27          & 66.92          & 64.93          & 52.99          & 54.48          & 49.07          \\
                              &                           & \cellcolor{gray!10}Ours                     & \cellcolor{gray!10}\textbf{49.25} & \cellcolor{gray!10}48.01          & \cellcolor{gray!10}\textbf{33.09} & \cellcolor{gray!10}\textbf{71.89} & \cellcolor{gray!10}\textbf{33.09} & \cellcolor{gray!10}\textbf{44.78} & \cellcolor{gray!10}\textbf{68.16} & \cellcolor{gray!10}\textbf{53.98} & \cellcolor{gray!10}\textbf{71.39} & \cellcolor{gray!10}\textbf{68.91} & \cellcolor{gray!10}\textbf{56.22} & \cellcolor{gray!10}\textbf{61.44} & \cellcolor{gray!10}\textbf{55.02} \\ \bottomrule
    \end{tabular}
}
% %\vspace{-0.15cm}
\label{Tab:sota}
% %\vspace{-0.3cm}
\end{table*}

\section{Main Experiments}
\label{sec:experiments}
\subsection{Dataset}
In our paper, we conduct experiments on AVMIT-C (action recognition) and AVE-C (event/object localization), in which the audio and video are highly correlated, allowing us to assess the benefits of audio in video TTA effectively. We also conduct experiments on the UCF101 and Kinetics-Sounds datasets, two commonly used action recognition datasets. Despite the low audio accuracy in these two datasets, our method remains effective.

\textbf{Audiovisual Moments in Time} (AVMIT)~\cite{joannou2023audiovisual} is a large-scale dataset of audiovisual action events. The dataset is collected from the Moments in Time dataset (MIT), comprising a total of 57,177 videos, each with a duration of 3 seconds. The test set consists of 960 videos, encompassing 16 common categories of human actions in daily life, with each category containing 60 videos.

\textbf{Audio-Visual Event} (AVE)~\cite{tian2018audio} is a subset of AudioSet~\cite{gemmeke2017audio}, that contains 4143 videos covering 28 event categories. The dataset includes audiovisual events from multiple domains, such as human activities, animal activities, music performances, and vehicle sounds. Each category contains at least 60 videos and at most 188 videos. The duration of the videos is at least 2 seconds, with approximately 66.4\% of the videos being 10 seconds long. The entire dataset is divided into three parts: 3339 videos are used for the training set and 402 videos for the test set.

\textbf{UCF101}~\cite{UCF101} is an action recognition dataset that provides 13,320 videos from 101 action categories. It is collected from YouTube and primarily includes five major categories of actions: interactions between people and objects, body movements, interactions between people, musical instrument performances, and sports activities. The official source provides three types of splits for dataset division. To follow ViTTA~\cite{vitta} and AME~\cite{AME}, we use the data of split 1, which consists of 9,537 and 3.783 videos for training and testing, respectively.

\textbf{Kinetics-Sounds}~\cite{arandjelovic2017look} is a subset of Kinetics dataset~\cite{kay2017kinetics}, which is a large-scale action recognition dataset sourced from YouTube. The Kinetics-Sounds dataset comprises 34 human action categories which have been chosen to be potentially manifested visually and aurally, with each category containing at least 400 videos. The dataset contains 15k training samples, 1.9k validation samples, and 1.9k test samples, and the duration of each video is approximately 10 seconds.

\subsection{Corruptions} We evaluate 12 types of corruption, which are introduced in~\cite{schiappa2022large,yi2021benchmarking} and related to changes in light intensity, weather conditions, and noise that exist in practical applications. These 12 corruptions are: Gaussian noise, pepper noise, salt noise, shot noise, contrast, impulse noise, rain, zoom blur, motion blur, jpeg compression, defocus blur and H265.abr.
% All our experiments are evaluated on the most severe corruption at level 5.
\dq{Previous TTA methods~\cite{TENT, SHOT, NORM,vitta}, including ViTTA, typically conduct experiments at severity level 5, as this represents the most challenging setting. To ensure a fair comparison with ViTTA, we also set the corruption severity level to 5.}

\begin{table*}[t!]
\caption{Comparison with state-of-the-art TTA methods on UCF-C and Kinetics-Sounds-C dataset using TANet w.r.t. accuracy (\%). * implement with official code.}
% \caption{Comparisons on datasets with low audio-visual correspondence. * implement with official code.}
\centering
\resizebox{\textwidth}{!}{
    \begin{tabular}{llcccccccccccc|c}
    \toprule
    \multirow{2}{*}{Dataset}         & \multirow{2}{*}{Methods} & \multicolumn{13}{c}{Corruptions}                                                                                                                                                                                           \\ \cline{3-15} 
                                     &                          & Gauss          & Pepper         & Salt           & Shot           & Contrast       & Impulse        & Rain           & Zoom           & Motion         & Jpeg           & Defocus        & H265.abr       & Avg.           \\ \hline
    \multirow{7}{*}{UCF101-C}        & Source                   & 17.92          & 23.66          & 7.85           & 72.48          & 76.04          & 17.16          & 37.51          & 54.51          & 83.40          & 62.68          & 81.44          & 81.58          & 51.35          \\
                                     & BN-Adapt\dq{~\cite{bn-adapt}}                 & 40.81          & 37.70          & 26.27          & 83.79          & 73.49          & 41.84          & 88.19          & 81.32          & 60.68          & 87.08          & 49.80          & 80.61          & 62.63          \\
                                     & TENT\dq{~\cite{TENT}}                     & 19.35          & 26.57          & 8.83           & 77.19          & 79.38          & 18.64          & 40.68          & 58.61          & 86.12          & 67.22          & 84.00          & 83.45          & 54.17          \\
                                     & SHOT\dq{~\cite{SHOT}}                     & 46.10          & 43.33          & 29.50          & 85.51          & 82.95          & 47.53          & 53.77          & 63.37          & 88.69          & 73.30          & 89.82          & 82.66          & 65.54          \\
                                     & NORM\dq{~\cite{NORM}}                     & 45.23          & 42.43          & 27.91          & 86.25          & 84.43          & 46.31          & 54.32          & 64.19          & 89.19          & 75.26          & 90.43          & 83.27          & 65.77          \\
                                     & ViTTA*\dq{~\cite{vitta}}                   & 69.78          & 64.18          & 44.68          & \textbf{92.04} & 87.64          & 71.53          & 68.35          & 80.53          & 91.59          & 86.97          & 92.62          & 84.53          & 77.87          \\
                                     &\cellcolor{gray!10}Ours                     &\cellcolor{gray!10}\textbf{71.66} & \cellcolor{gray!10}\textbf{66.46} & \cellcolor{gray!10}\textbf{47.40} & \cellcolor{gray!10}91.81          & \cellcolor{gray!10}\textbf{88.07} & \cellcolor{gray!10}\textbf{72.72} & \cellcolor{gray!10}\textbf{71.34} & \cellcolor{gray!10}\textbf{81.63} & \cellcolor{gray!10}\textbf{91.62} & \cellcolor{gray!10}\textbf{87.30} & \cellcolor{gray!10}\textbf{92.68} & \cellcolor{gray!10}\textbf{84.31} & \cellcolor{gray!10}\textbf{78.92} \\ \hline
    \multirow{7}{*}{Kinetic-sound-C} & Source                   & 46.51          & 48.27          & 30.76          & 77.60          & 34.62          & 48.33          & 71.65          & 59.50          & 79.49          & 77.47          & 63.68          & 50.88          & 57.40          \\
                                     & BN-Adapt\dq{~\cite{bn-adapt}}                 & 50.36          & 49.77          & 34.62          & 76.42          & 41.48          & 49.77          & 77.14          & 63.36          & 78.18          & 74.66          & 59.37          & 42.33          & 58.12          \\
                                     & TENT\dq{~\cite{TENT}}                     & 51.27          & 52.25          & 36.06          & 77.86          & 41.80          & 51.86          & 79.49          & 68.00          & 80.01          & 77.20          & 61.79          & 43.44          & 60.09          \\
                                     & SHOT\dq{~\cite{SHOT}}                     & 55.52          & 54.28          & 38.34          & 78.25          & 45.85          & 55.26          & 79.88          & 69.89          & \textbf{80.08} & 77.47          & 63.49          & 45.00          & 61.94          \\
                                     & NORM\dq{~\cite{NORM}}                     & 51.34          & 51.86          & 36.12          & 77.60          & 41.28          & 51.40          & 78.84          & 67.21          & 79.69          & 76.81          & 61.53          & 43.37          & 59.75          \\
                                     & ViTTA*\dq{~\cite{vitta}}                   & 69.17          & 65.97          & 47.68          & 81.25          & 60.68          & 68.91          & 80.67          & 68.84          & 79.43          & 78.84          & 68.84          & 49.45          & 68.31          \\
                                     & \cellcolor{gray!10}Ours                     & \cellcolor{gray!10}\textbf{71.20} &\cellcolor{gray!10}\textbf{68.71} &\cellcolor{gray!10}\textbf{53.76} &\cellcolor{gray!10}\textbf{82.17} &\cellcolor{gray!10}\textbf{63.68} &\cellcolor{gray!10}\textbf{70.22} &\cellcolor{gray!10}\textbf{81.65} &\cellcolor{gray!10}\textbf{70.08} &\cellcolor{gray!10}79.29          &\cellcolor{gray!10}\textbf{79.69} &\cellcolor{gray!10}\textbf{71.98} &\cellcolor{gray!10}\textbf{51.67} &\cellcolor{gray!10}\textbf{70.34} \\ \bottomrule
    \end{tabular}
  }
  \label{tab:ucf}
\end{table*}

\begin{table*}[t]
\centering
\caption{\dq{Effectiveness of audio information for video TTA using TANet w.r.t accuracy(\%).}} 
\resizebox{\textwidth}{!}
{
\begin{tabular}{>{\color{black}}c<{} >{\color{black}}l<{} *{13}{>{\color{black}}c<{}}} % 自动标蓝列格式
\toprule
\multicolumn{1}{>{\color{black}}c<{}}{} & \multicolumn{1}{>{\color{black}}c<{}}{} & \multicolumn{13}{>{\color{black}}c<{}}{Corruptions} \\ 
\cline{3-15} 
\multicolumn{1}{>{\color{black}}c<{}}{\multirow{-2}{*}{Dataset}} & \multicolumn{1}{>{\color{black}}c<{}}{\multirow{-2}{*}{Method}} & Gauss & Pepper & Salt  & Shot  & Contrast & Impulse & Rain  & Zoom  & Motion & Jpeg  & Defocus & H265.abr & Avg. \\ 
\hline
\multirow{2}{*}{AVMIT-C} & w/o audio               & 52.71 & 49.58 & 38.65 & 77.19 & 36.77 & 51.88 & 71.15 & 56.35 & 75.10 & 77.19 & 65.42 & 60.73 & 59.39        \\
                         & w/ audio                & 72.40 & 63.85  & 65.73 & 84.06 & 63.65    & 71.88   & 80.94 & 72.81 & 83.75  & 86.56 & 78.85   & 72.92    & \textbf{74.78} \\ \hline
\multirow{2}{*}{AVE-C}   & w/o audio               & 53.48 & 52.74 & 37.81 & 69.65 & 35.32 & 53.73 & 64.93 & 48.51 & 67.91 & 68.16 & 51.00 & 53.73 & 54.75         \\
                         & w/ audio                & 62.44 & 63.68  & 51.74 & 74.63 & 47.51    & 59.70   & 70.65 & 62.44 & 70.40  & 70.15 & 61.94   & 58.21    & \textbf{62.79} \\ 

\bottomrule
\end{tabular}

}
\label{audio_gain}
\end{table*}

\subsection{Model Architecture}
We evaluate our approach on two different model architectures: TANet-R50~\cite{liu2021tam} and TSM-R50~\cite{Lin_Gan_Han_2020}, both of which are based on ResNet50~\cite{he2016deep} for video classification. We use TANet as the default model unless otherwise specified.

\subsection{Implementation Details}
We divide each video into 16 segments equally. A frame is randomly selected in the first segment, and 16 frames are sampled equidistantly. We set the batch size to 1 and perform online TTA. \dq{We use GPT-4~\cite{gpt4} as the default LLM.}

\textbf{AVMIT-C dataset:} The learning rate for experiments using TANet as the backbone is set at 5e-6, while for those using TSM as the backbone, it is set at 1e-6. The maximum number of adaptation cycles, $\tau$, is set to 8 on both backbones.

\textbf{AVE-C dataset:} The learning rate for experiments with TANet as the backbone is set at 1e-5, and for those with TSM as the backbone, it is maintained at 1e-6. $\tau$ is set to 8 and 4 respectively on two backbones, \dq{as the performance tends to stabilize when $\tau$ exceeds 4 on TSM.}

\textbf{UCF101-C and Kinetics-Sounds-C datasets:} We follow the learning rates provided by ViTTA for experiments. Using TANet as the backbone, we set the learning rates to 5e-5, with $\tau$ values of 3 and 2, respectively.

\begin{figure*}[!t]
    \centering 
    \includegraphics[width=0.8\textwidth]{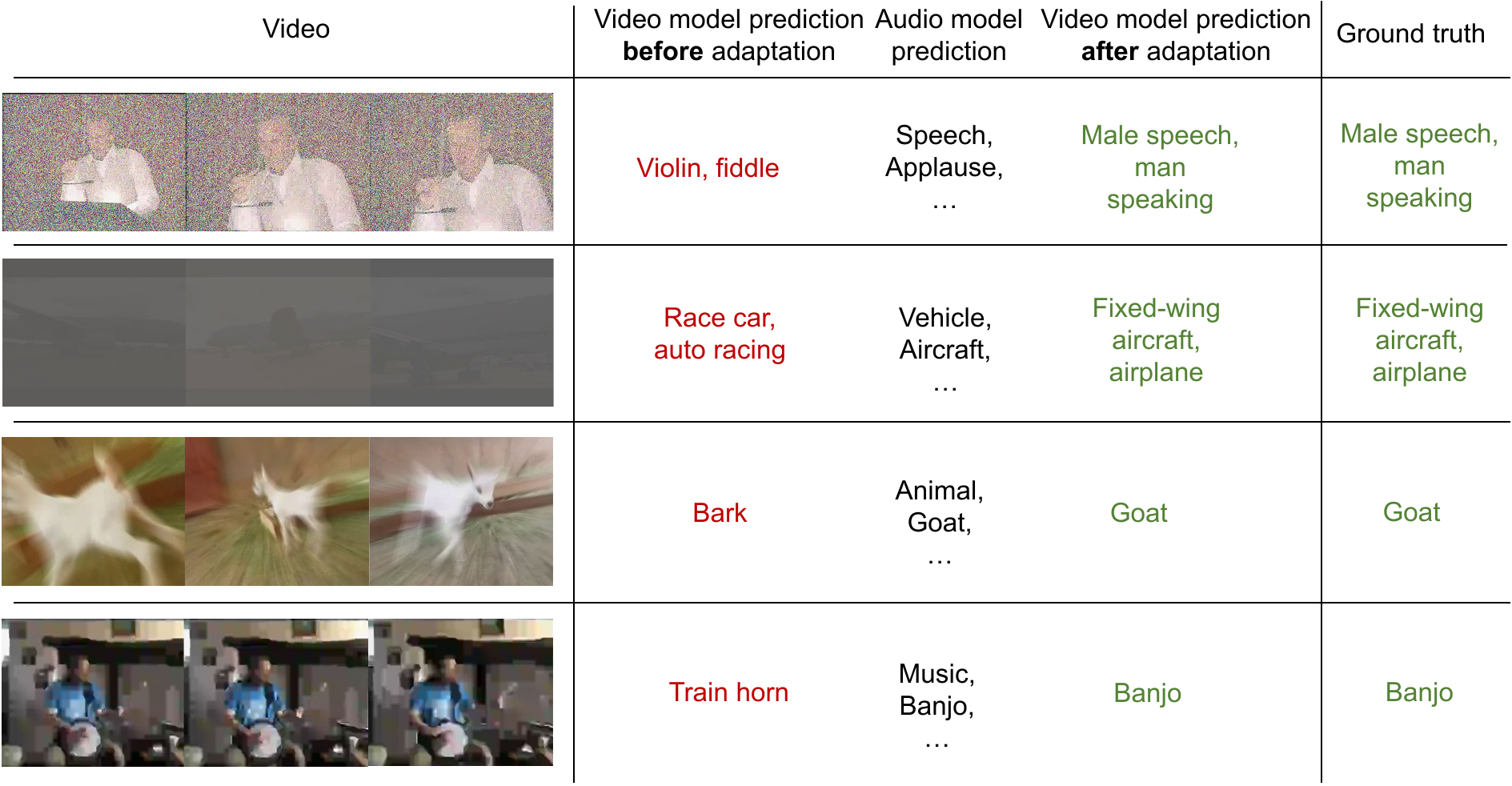}
    \caption{\dq{Qualitative analysis of the results before and after adaptation using audio information.}} 
    \label{fig:visual_result}
\end{figure*}

\subsection{Comparisons with State-of-the-Art}

% \noindent \textbf{Compared methods.}
\textbf{Results on AVMIT-C and AVE-C.}
% We compare our proposed method with the state-of-the-art video TTA method (\ie, ViTTA~\cite{vitta}) and three image-based TTA methods are selected, namely, TENT, SHOT, and NORM. We also implement a BN-Adapt baseline,
\dq{We compare our proposed method with the state-of-the-art video TTA method (\ie, ViTTA~\cite{vitta}) and image-based TTA methods. We also implement a BN-Adapt baseline~\cite{bn-adapt},}
which updates the running estimations of mean and variance statistics in batch normalization layers during adaptation. For fair comparisons, we use the same backbone feature extractors. The adaptation results on AVMIT-C and AVE-C are reported in Table~\ref{Tab:sota}. It shows that image-based TTA methods achieve a slight improvement over those without adaptation. ViTTA is generally superior to those image-based approaches. However, there is still a noticeable gap when compared to our method. Our method consistently surpasses the aforementioned comparative methods across twelve different noise scenarios. This is due to our approach's ability to consider and utilize the latent audio information in videos, and its adaptive capacity to flexible adaptation cycle for each case, thereby achieving enhanced adaptive effects.

% \dq{We further present experimental results for different corruption levels in Table~\ref{tab:different_level}. These results show that our method consistently outperforms ViTTA across all corruption levels, and as the severity level increases, the performance improvement achieved by our method becomes more pronounced. This indicates that audio is beneficial across all corruption levels, with the aid of audio becoming more significant as the corruption severity of the video increases.}

\textbf{Results on UCF101-C and Kinetics-Sounds-C.}
% \subsection{Can Our Method Work with Low Audio Accuracy?}
% Significant improvements on datasets with high audio-video relevance, such as AVMIT-C and AVE-C, demonstrate the effectiveness of utilizing audio in video TTA. In addition, 
We present comparisons using the UCF101-C and Kinetics-Sounds-C datasets in Table~\ref{tab:ucf}. On the Kinetics-Sounds-C dataset, our method improves the accuracy from 68.31\% to 70.31\%. For the UCF101 dataset, only 1944 samples (51.5\%) contain audio, and even for those with audio, the audio-video correlation is relatively weak, as evidenced by the low pseudo-label accuracy of 9.67\% on UCF101-C. Despite these challenges, our method still improves TTA performance from 77.87\% to 78.92\%, as shown in Table~\ref{tab:ucf}. This improvement can primarily be attributed to the inclusion of two additional loss terms in Eqn. \ref{Eq:total_loss}, which enables the use of both visual and audio cues for TTA. 

%这个数据集上的提升相对于AVMIT-C and AVE-C没那么大，可能原因是audio-video relevance相对weaker，As claimed in~\cite{arandjelovic2017look}, \textit{``although the Kinetics-Sounds dataset is fairly clean by construction, it still contains considerable noise, e.g. human voices often masks the sound of interest and many videos contain soundtracks that are completely unrelated to the visual content''}. 
% To explore the upper bound of performance under the same hyperparameter settings, we substituted the pseudo labels with ground truth labels, achieving an accuracy of 72.93\%. The proximity of our method's performance to this upper bound underscores its effectiveness in leveraging both audio features and pseudo labels generated by the LLM.

\section{Ablation Study}
In this section, we conduct in-depth ablation experiments on four corruptions (unless otherwise specified), including Gauss, Pepper, Salt, and Shot, to understand the effectiveness of each proposed component.

\subsection{Does Audio-Assisted Supervision Help TTA?}
Compared with existing TTA methods, our primary distinction lies in our utilization of audio information. For fair comparisons, we conduct experiments to contrast the scenarios with and without the use of audio (\ie, the $L_{cls}$ in Eqn. (\ref{cls})), while maintaining a flexible adaptation cycle in both cases. \dq{As illustrated in Table~\ref{audio_gain}, on the AVMIT-C and AVE-C datasets, the incorporation of audio consistently results in significant performance improvements under all 12 types of corruption, where enhancements reach approximately 15\% and 8\%, respectively.} This underscores the effectiveness of our method stemming from the efficient use of audio. Our approach offers a novel perspective for enhancing TTA performance by leveraging multimodal information within videos.

\begin{table*}[h]
\centering
\caption{\dq{Ablation of flexible adaptation cycle on AVE-C dataset using TANet w.r.t accuracy(\%).}} 
\resizebox{\textwidth}{!}
{
\begin{tabular}{>{\color{black}}l<{} >{\color{black}}l<{} *{13}{>{\color{black}}c<{}}} % 自动标蓝列格式
\toprule
Method                 & Setting        & Gauss & Pepper & Salt  & Shot  & Contrast & Impulse & Rain  & Zoom  & Motion & Jpeg  & Defocus & H265.abr & Avg.           \\ \hline
\multirow{3}{*}{ViTTA} & No cycle       & 51.00 & 53.23  & 36.07 & 69.65 & 36.07    & 51.54   & 64.43 & 48.01 & 69.65  & 67.66 & 51.24   & 53.73    & 54.36          \\
                       & Fixed cycle    & 52.24 & 51.99  & 38.06 & 68.16 & 35.32    & 53.23   & 64.93 & 49.25 & 67.91  & 68.16 & 50.00   & 53.73    & 54.42          \\
                       & Flexible cycle & 53.48 & 52.74 & 37.81 & 69.65 & 35.32 & 53.73 & 64.93 & 48.51 & 67.91 & 68.16 & 51.00 & 53.73 & \textbf{54.75} \\ \hline
\multirow{3}{*}{Ours}  & No cycle       & 52.99 & 53.73  & 42.29 & 69.15 & 39.55    & 54.48   & 66.42 & 51.49 & 68.41  & 69.65 & 55.22   & 53.23    & 56.38          \\
                       & Fixed cycle    & 57.21 & 59.95  & 49.25 & 69.90 & 43.78    & 61.19   & 69.40 & 61.94 & 71.14  & 67.91 & 59.70   & 60.20    & 60.97          \\
                       & Flexible cycle & 62.44 & 63.68  & 51.74 & 74.63 & 47.51    & 59.70   & 70.65 & 62.44 & 70.40  & 70.15 & 61.94   & 58.21    & \textbf{62.79} \\ \bottomrule
\end{tabular}
}
\label{tab:cycle}
\end{table*}

\dq{\textbf{More qualitative analysis.}}
\dq{We further conduct a qualitative analysis on the AVE-C dataset to understand the effectiveness of the proposed method in Figure~\ref{fig:visual_result}. For example, in the first row, corruption significantly interferes with the visual information, leaving only a vague outline of the human body with no detailed motion features visible. In this scenario, the video model incorrectly predicts the action as ``Violin, fiddle.'' However, by leveraging the audio information from the video, we use the audio model to predict the categories ``Speech'' and ``Applause'', which are then mapped to the video label "Male speech, man speaking" through the LLM. This process adapts the model, leading to a correct prediction by the video model. Other examples further demonstrate that audio information can provide reliable supervisory signals, especially when the video data is corrupted, thereby improving test-time adaptation performance.}

\subsection{Do We Need Adaptation Cycle?}\label{Sec:reuse}
We aim to investigate whether it is feasible to utilize pseudo labels obtained by our method for multiple adaptations of the model. 
We incorporate a baseline, which involves setting $\alpha = 0$ to remove the $\mathcal{L}_{cls}$ term from Eqn. (\ref{Eq:total_loss}) (that is, not utilizing audio-assisted pseudo labels). This equals combining ViTTA~\cite{vitta} with multiple adaptations. 
Specifically, during the TTA process, each video sample from the online streaming input is tested after updating the model using Eqn. (\ref{Eq:total_loss}) for \( t \in \{1,2, \ldots, 7\} \) iterations. In each experiment, the number of adaptations per sample remains consistent. 

\begin{figure}[!t]
    \centering 
    \includegraphics[width=\linewidth]{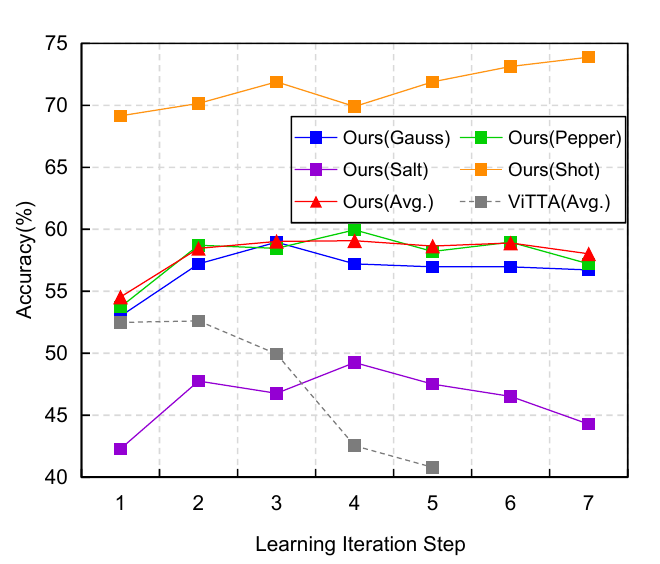}
    % %\vspace{-0.4cm}
    \caption{Ablation on TTA's learning iteration steps (\ie, adaptation cycles) on TANet. Without using audio, ViTTA's accuracy drops significantly with more steps. The optimal step under different corruption types varies.}
    % \caption{Ablation on TTA's learning iteration steps (\ie, adaptation cycles) on TANet.}
    \label{Fig4}
    % %\vspace{-0.4cm}
\end{figure}

The results in Fig~\ref{Fig4} indicate that multiple adaptations are more effective than a single adaptation, suggesting the necessity of an adaptation cycle. We found that directly applying adaptation cycles to ViTTA without considering audio information is not effective. As the number of adaptations increases, the performance consistently decreases. Therefore, in this case, we only present the results of five cycles of ViTTA. This indicates that adaptation cycles need to be integrated with effective supervision. Our method offers a feasible approach for extracting supervisory information during TTA. Concurrently, we also discovered that the optimal number of repetitions varies across different settings (different corruptions). This finding motivated us to design a flexible adaptation cycle.

\begin{table}[t]
\caption{\dq{Ablation of flexible cycle conditions in Eqn.~(\ref{Eq:condition}) using TANet w.r.t averaged accuracy across 12 types of corruption.}}
\centering
\scriptsize
\setlength{\tabcolsep}{3pt}
\begin{tabular}{>{\color{black}}l<{} *{4}{>{\color{black}}c<{}}} % 自动标蓝列格式
\toprule
Dataset & $\mathcal{L}_{align}^{(t)}\small{<}\mathcal{L}_{align}^{(t-1)}$  & $\mathcal{L}_{cons}^{(t)}\small{<}\mathcal{L}_{cons}^{(t-1)}$  & $\hat{y}_i^{(t)} \small{\neq} \hat{y}_j^{(t)}$  & Avg. Acc. (\%) \\ 
\midrule
\multirow{5}{*}{AVMIT-C}   
&           &         &          & 71.91                        \\
& \checkmark    & \checkmark  &          & 68.50                        \\
& \checkmark    &         & \checkmark   & 72.07                        \\
&           & \checkmark  & \checkmark   & 73.16                        \\
& \checkmark    & \checkmark  & \checkmark   & \textbf{74.78}               \\ 
\midrule
\multirow{5}{*}{AVE-C}     
&           &         &          & 60.97                        \\
& \checkmark    & \checkmark  &          & 59.10                        \\
& \checkmark    &         & \checkmark   & 60.90                        \\
&           & \checkmark  & \checkmark   & 61.82                        \\
& \checkmark    & \checkmark  & \checkmark   & \textbf{62.79}               \\ 
\bottomrule
\end{tabular}
\label{tab:condition}
\end{table}

\begin{table}[t]
\centering
\caption{\dq{Results with partially available pseudo labels using TANet w.r.t average accuracy(\%) across 12 types of corruption.}} 
\resizebox{\columnwidth}{!}
{
\begin{tabular}{>{\color{black}}l<{} *{7}{>{\color{black}}c<{}}}
\toprule
\multirow{2}{*}{Dataset} & \multirow{2}{*}{Source} & \multirow{2}{*}{ViTTA} & \multicolumn{5}{c}{\dq{Ratio of available pseudo label($m$\%)}} \\ \cmidrule{4-8} 
                         &                         &                        & 20\%    & 40\%    & 60\%   & 80\%   & 100\%           \\ \midrule
AVMIT-C                  & 51.07                   & 59.76                  & \textbf{60.12}   & 61.03   & 64.37  & 69.92  & 74.78  \\
AVE-C                    & 44.05                   & 54.36                  & \textbf{55.06}   & 55.83   & 56.74  & 58.62  & 62.79  \\ \bottomrule
\end{tabular}
}
\label{tab:drop}
\end{table}

\subsection{Does Flexible Adaptation Cycle Boost TTA?}

To adaptively adjust the adaption steps for each sample, we propose a flexible adaptation cycle, wherein the criteria consider three conditions (Eqn. (\ref{Eq:condition})): \(\mathcal{L}_{align}^{(t)} < \mathcal{L}_{align}^{(t-1)}\), \(\mathcal{L}_{cons}^{(t)} < \mathcal{L}_{cons}^{(t-1)}\), and \(\hat{y}_i^{(t)} \neq \hat{y}_j^{(t)}\). Our ablation experiments consider combinations of two conditions and the use of all three conditions and compare them with a scenario where no selection criteria are applied.
For fair comparisons, in the absence of selection criteria, we also employ a strategy of multiple adaptation cycles and compare the best results obtained from the optimal number of cycles. The results in Table \ref{tab:condition} indicate that using only the first two conditions as constraints leads to nearly unchanged or even slightly decreased accuracy. However, when condition 3, which assesses prediction consistency, is considered in combination with either condition 1 or 2, there is a significant performance improvement. 
% Furthermore, combining condition 3 with other conditions also results in substantial improvements. 
This demonstrates the vital role of cross-view consistency in determining the adaptation cycle. When all conditions are combined, the results are optimal, suggesting that the flexible adaptation cycle strategy can effectively select the number of sample repetitions, thereby achieving better TTA performance, as also verified in Table \ref{tab:cycle}.

% In practical applications, low audio accuracy is frequently observed due to the background music or noise in videos, which subsequently leads to reduced accuracy in pseudo video labels. For instance, the UCF101-C dataset, compared to AVMIT-C, is more prone to audio-visual discrepancies, resulting in a pseudo video label accuracy of only 9.67\%. From Table~\ref{tab:video_audio}, even under these challenging conditions, our method still enhances the TTA performance from 68.30\% to 69.33\%. This improvement can primarily be attributed to the incorporation of two additional loss terms in Equation (9) within our proposed flexible adaptation cycle, which utilizes visual cues in addition to audio for TTA. The exploration of methods to further enhance TTA performance with low audio accuracy presents an intriguing avenue for research, which we intend to pursue in our future work.

\subsection{When Pseudo Video Labels Are Not Readily Available?}
In real scenarios, there may be instances where the audio corresponding to some videos is unobtainable, or there could be issues with network latency (\eg, when utilizing LLMs through API calls). These factors can ultimately lead to the inability to promptly acquire pseudo labels. Therefore, we design an experiment by keeping each video's pseudo label with a probability $m$\%, that is, simulating the situation where the entire dataset has only a probability of $m$\% that a pseudo video label can be obtained. From Table~\ref{tab:drop}, even with only 20\% of the pseudo video labels available in AVMIT-C and AVE-C, our method significantly outperforms ViTTA, demonstrating its practical applicability.

% \begin{table}[t]
% \caption{Comparison with state-of-the-art TTA methods on UCF-C and Kinetics-Sounds-C dataset w.r.t. accuracy (\%). * implement with official code.}
% % \caption{Comparisons on datasets with low audio-visual correspondence. * implement with official code.}
% \centering
% \resizebox{.8\linewidth}{!}
% {
%     \begin{tabular}{lcc}
%     \toprule
%     Methods          & UCF101-C       & Kinetics-Sounds-C \\ \midrule
%     Source           & 51.35          & 57.40           \\
%     BN-Adapt         & 62.63          & 58.12           \\
%     TENT             & 54.17          & 60.09           \\
%     SHOT             & 65.54          & 61.94           \\
%     NORM             & 65.77          & 59.75           \\
%     % ViTTA            & 78.33          & -           \\
%     ViTTA* & 77.87          & 68.31              \\
%     Ours             & \textbf{78.92} & \textbf{70.34}                  \\ \bottomrule
%     \end{tabular}
%   }
%   \label{tab:ucf}
% \end{table}

\begin{table*}[!t]
\caption{Results of label mapping on AVMIT-C dataset. Based on the audio labels and their probability(shown in parentheses) obtained from the audio classifier, LLM is capable of delivering plausible predictions.}
  \centering
  \renewcommand\arraystretch{1}%行高
  % \tabcolsep 8pt
  % \footnotesize
  \setlength{\tabcolsep}{4pt}  % 设置列间距
  \large
  
  \resizebox{\textwidth}{!}{
      % \begin{tabular}{@{}l@{\hspace{2em}}l@{}}
\begin{tabular}{lllllc}
    \toprule
    \multicolumn{5}{c}{Top-5 audio predictions}                                                                                                             & \multirow{2}{*}{\begin{tabular}[c]{@{}c@{}}LLM\\ prediction\end{tabular}} \\ \cline{1-5}
    Category 1                   & Category 2                      & Category 3                      & Category 4                    & Category 5                     &                                                                           \\ \hline
    Dog(0.63)                 & Animal(0.61)                 & Domestic animals, pets(0.52) & Bark(0.44)                 & Bow-wow(0.44)               & barking                                                                   \\
    \rowcolor{gray!10}Vehicle(0.60)             & Lawn mower(0.20)             & Propeller, airscrew(0.11)    & Motorboat, speedboat(0.06) & Aircraft(0.06)              & mowing                                                                    \\
    Music(0.47)               & Drum(0.45)                   & Percussion(0.33)             & Drum kit(0.25)             & Drum roll(0.22)             & drumming                                                                  \\
    \rowcolor{gray!10}Wood(0.95)                & Rub(0.90)                    & Sawing(0.68)                 & Sanding(0.26)              & Filing (rasp)(0.02)         & sanding                                                                   \\
    Sizzle(0.19)              & Frying (food)(0.14)          & Spray(0.06)                  & Inside, small room(0.05)   & Speech(0.04)                & frying                                                                    \\
    Baby laughter(0.50)       & Laughter(0.41)               & Belly laugh(0.16)            & Giggle(0.09)               & Babbling(0.08)              & giggling                                                                  \\
    \rowcolor{gray!10}Applause(0.56)            & Rain on surface(0.19)        & Raindrop(0.14)               & Rain(0.10)                 & Sound effect(0.09)          & raining                                                                   \\
    Vehicle(0.57)             & Car(0.13)                    & Heavy engine(0.07)           & Truck(0.06)                & Car passing by(0.06)        & mowing                                                                    \\
    \rowcolor{gray!10}Clicking(0.18)            & Speech(0.06)                 & Stomach rumble(0.04)         & Inside, small room(0.02)   & Sound effect(0.02)          & tapping                                                                   \\
    Boat, Water vehicle(0.15) & Gurgling(0.09)               & Water(0.07)                  & Snort(0.06)                & Rowboat, canoe, kayak(0.06) & diving                                                                    \\
    \rowcolor{gray!10}Animal(0.84)              & Domestic animals, pets(0.79) & Dog(0.75)                    & Howl(0.73)                 & Canidae, dogs, wolves(0.33) & howling                                                                   \\ \bottomrule
    \end{tabular}
  }
  \label{table:label_mapping}

\end{table*}

\begin{table}[t]
\caption{Comparisons on runtime (seconds/video). $\tau$: the maximum number of adaptation cycles for a sample, $s$: the number of delayed samples in the delayed update strategy. We report the average accuracy under four types of corruption: Gauss, Pepper, Salt, and Shot.}
\centering
\resizebox{\linewidth}{!}
{
    \begin{tabular}{lccccc}
    \toprule
    \multirow{2}{*}{AVMIT-C}     &\multicolumn{2}{c}{\textbf{Parallel} Processing}                   & \multirow{2}{*}{Runtime}   & \multirow{2}{*}{FPS} & \multirow{2}{*}{Accuracy(\%)}\\ \cmidrule{2-3}
                                & Video    & Audio                  &       &       \\ \midrule 
    ViTTA               & 0.097s & N/A    & 0.097s & 164.9 & 54.53 \\
    Ours($\tau$=1, $s$=0)   & 0.097s & 0.260s & 0.321s & 49.8 & 59.22 \\
    Ours($\tau$=2, $s$=0)   & 0.161s & 0.260s & 0.385s & 41.6 & 60.42 \\
    Ours($\tau$=4, $s$=0)   & 0.236s & 0.260s & 0.460s & 34.8 & 66.40 \\
    Ours($\tau$=8, $s$=0)      & 0.301s & 0.260s & 0.525s & \textbf{30.5} & \textbf{71.51} \\ \midrule
    Ours($\tau$=1, $s$=1)   & 0.097s & 0.260s & 0.306s & 52.3 & 59.38 \\
    Ours($\tau$=1, $s$=2)   & 0.097s & 0.130s & 0.194s & 82.5 & 59.38 \\
    Ours($\tau$=1, $s$=4)   & 0.097s & 0.065s & 0.138s & 115.9 & 58.91 \\
    Ours($\tau$=1, $s$=8)   & 0.097s & 0.033s & 0.110s & 145.5 & 58.78 \\
    Ours($\tau$=1, $s$=16)   & 0.097s & 0.016s & 0.097s & \textbf{164.9} & \textbf{58.46} \\ \midrule
    Ours($\tau$=8, $s$=1)   & 0.301s & 0.260s & 0.510s & 31.4 & 68.05 \\
    Ours($\tau$=8, $s$=2)   & 0.301s & 0.130s & 0.398s & 40.2 & 67.32 \\
    Ours($\tau$=8, $s$=4)   & 0.301s & 0.065s & 0.342s & 46.8 & 67.06 \\
    Ours($\tau$=8, $s$=8)   & 0.301s & 0.033s & 0.314s & 51.0 & 63.18 \\
    Ours($\tau$=8, $s$=16)   & 0.301s & 0.016s & 0.301s & 53.2 & 60.55 \\ 
    \bottomrule  
    \end{tabular}
}
\label{tab:runtime}
\end{table}

\subsection{Is our audio-assisted TTA method efficient?}
We conduct runtime comparisons with TANet backbone using a single A800 GPU in Table \ref{tab:runtime}, revealing several key observations: \textbf{1)} Our approach operates in real-time, compatible with the practical video frame rate of 30 FPS. \textbf{2)} Our method is flexible, allowing for adjusting $\tau$, the maximum number of adaptation cycles, to accommodate varying FPS based on specific application requirements. To facilitate higher FPS, we propose a simple delayed update strategy. Specifically, given a continuous stream of test data, our model updates (incorporating TTA) only after processing a predefined number of samples, denoted as \(s\). Within this period, we await the outcomes of pseudo video labels before applying model updates. Our approach consistently surpasses ViTTA across various \(s\) values. Notably, at \(\tau=1\) and \(s=16\), our method significantly outperforms ViTTA (58.46\% vs 54.53\%) with the same 164.9 FPS.

\begin{table}[t]
\caption{Results on the UCF101-C dataset with optional proposed strategies using TANet w.r.t. accuracy (\%). The results demonstrate the effectiveness of the proposed strategies.}
  \centering
  \resizebox{\columnwidth}{!}{
    \begin{tabular}{lccccc}
    \toprule
    \multirow{2}{*}{Methods} & \multicolumn{5}{c}{Corruptions} \\ \cmidrule{2-6}
                        & Gauss  & Pepper & Salt  & Shot  & Avg.  \\ \midrule
    ViTTA               & 71.37 & 64.55 & 45.84 & 91.44 & 68.30 \\ %paper
    % ViTTA & 70.94 & 63.97 & 44.7  & 91.57 & 67.80 \\ %复现
    Ours                & 71.66 & 66.46 & 47.40 & 91.81 & 69.33 \\
    Ours+ensemble       & 71.77 & 66.86 & 48.32 & 91.80 & 69.69 \\
    Ours+select         & 72.27 & 67.10  & 48.54 & 91.86 & 69.94 \\
    Ours+select+ensemble& 72.40 & 67.47 & 49.20 & 91.86 & \textbf{70.23} \\ 
    \bottomrule
    \end{tabular}%
    }
  \label{tab:addlabel}%
  \vspace{-0.4cm}
\end{table}%

\label{sec:further}

\subsection{Analysis of LLM's pseudo label predictions}
\label{sec:label_analysis}
% 我们选取了音视频相关性较低的数据集UCF101来对LLM预测结果准确性进行分析。如图X所示，我们从UCF101中选取一个子集（约1k个视频），分析了导致LLM预测不准确的主要原因：
% We select a dataset with low audio and video correlation to analyze the accuracy of pseudo label predictions. 
Upon inputting this specified prompt into LLMs, we guide them to execute the process of label mapping. The results of label mapping on the AVMIT-C dataset are presented in Table \ref{table:label_mapping}. The first column of the table enumerates the top-5 audio labels along with their respective probabilities, marked by a colon separating the audio label and its probability. The second column demonstrates the most relevant video label chosen by LLMs from the video label space, based on the provided input. For instance, considering the first sample in the table, audio labels such as ``Dog'', ``Animal'', ``Bark'', and others, each with a probability, are fed into LLMs. Utilizing these labels, LLMs deduce the video label to be ``barking'', a conclusion that appears highly logical.

\begin{figure*}[t]
    \centering 
    \includegraphics[width=0.9\textwidth]{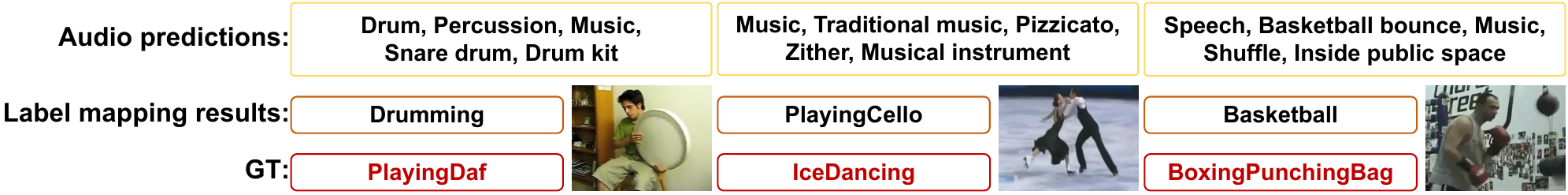}
    % \vspace{3pt}
    % \centering
    \caption{Three cases of inaccurate LLM pseudo labels.} 
    \label{failure_case}
    \vspace{-10pt}
\end{figure*}

We analyze a subset of UCF101 ($\sim$1k samples) and identify three types of error in LLM's predictions, as shown in Fig.~\ref{failure_case}:
% 1) Different objects producing similar sounds;如左侧例子所示，daf也是鼓类的一种，产生的音频相似性极高，因此仅靠音频难以区分；
\textbf{1)} Different objects produce similar sounds. From the left example, the daf is also a type of drum. The audio produced is very similar, so it is difficult to distinguish it by audio alone.
% 2) Mismatch between audio and vision;如中间例子所示，冰舞表演的背景音乐为cello演奏，因此在音频中没有能体现冰舞的信息；
\textbf{2)} Mismatch between audio and vision. As shown in the middle example, the background music of the ``IceDancing'' is played by cello, there is no information that can reflect the ice dance in the audio.
% 3) Similarity in ambient scene sounds. 如右边例子所示，boxing和basketball对应的场所均为运动馆，均包含人们交谈欢呼以及走动的声音。因此在场所背景音频相似的情况下难以区分具体的动作。
\textbf{3)} Similarity in ambient scene sounds. From the right example, the places corresponding to boxing and basketball are both sports halls. They both contain the sounds of people talking, cheering, and moving around. Therefore, it is difficult to distinguish specific actions when the ambient scene sound is similar. Despite these factors potentially leading to low accuracy of pseudo labels (\eg, 9.67\% on UCF101), we still achieve better performance across 15 corruptions, \ie, 78.92\% (ours) vs. 77.87\% (ViTTA).

% We analyze a subset of UCF101 ($\sim$1k samples) and identify three types of error in LLM's predictions: 
% 1) Different objects producing similar sounds; 2) Mismatch between audio and vision; 3) Similarity in ambient scene sounds. 

% Despite these factors potentially leading to low accuracy of pseudo labels (\eg, 9.67\% on UCF101), we still achieve better performance, \ie, 69.33\% (ours) \vs~68.30\% (ViTTA).
% % our approach still achieves an improvement over the baseline in ViTTA, \ie, 69.33\% vs 67.80\% in Table~\ref{tab:video_audio}.

\subsection{Optional label filtering strategies}
\label{sec:filter_labels}
Consider a realistic situation, when the audio and video correlation is low, the pseudo labels may be unreliable. Therefore, based on our method, we additionally design two optional strategies and can choose whether to use or discard the pseudo label to obtain better performance.
% （策略1：select）
% 对于音画相关性较高的数据集，如：AVE、AVMIT，我们认为通过LLM得到的label是相对比较可靠的，则无需对label做进一步的筛选。而有些数据集音画相关性较低，如：UCF101、K400，我们认为这种情况下根据音频得到的label与视频动作的相关性较低，需要对LLM得到的label做进一步的筛选。我们根据pseudo label在video的预测中的排序进行判断，若位于前top-k，则认为该label是可靠的，即可使用；若不在top-k中，则认为该label与视频动作类别差距较大，即不进行使用。
\textbf{1) Select strategy}: 
We assess the reliability of pseudo-labels by comparing them with video predictions. A pseudo-label is deemed trustworthy and therefore applicable if it ranks within the top-$k$ predictions for a given video. Otherwise, such a label will not be used in the TTA process. 
% For the datasets with low correlation between audio and video, such as UCF101 and K400, the label obtained based on the audio has a low accuracy. So further filtering is required. We determine the reliability of pseudo label based on its ranking in the prediction of the video. If it is in the top-$k$ of video prediction, it is considered that the label is reliable and can be used. Otherwise, it is considered that the label is far behind the video action and it will not be used.
%（策略2：ensemble）
% 在inference阶段，我们可以使用常见的video预测结果作为最终预测，也可以考虑将audio信息进行使用。audio除了可以为我们TTA过程提供监督信息以外，还可以考虑将pseudo video label （PVL）和video model的预测进行融合作为最终的预测结果。但由于pseudo video label是由LLM直接得到的one hot label，不包含对各个类的预测权重信息，而video model predict的是soft label，包含每类概率。因此我们不能简单地将两者进行加权求和或者投票取最大类别的方式来进行融合。因此，我们在实验中探索了一种简单的audio-visual融合策略：我们认为当video model预测可信度低且PVL更加可靠时，可以使用PVL作为最终预测结果，其余情况均使用video model的结果作为最终预测结果。具体地，我们用video model预测类别对应的概率的值来判断其预测置信度，当其小于阈值$\theta$时说明预测置信度低。关于PVL是否可靠的，我们沿用与前面所讲筛选策略的方法进行判断，即当PVL位于video预测排序的前top-k时，则认为PVL是可靠的。总的来说，当video model预测类别置信度低，且PVL更加可靠时，我们用PVL作为最终结果，否则用video的预测作为最终结果。
\textbf{2) Ensemble strategy}: When the confidence of the predicted class (outputted by the video model) is below a threshold $\theta$ and the pseudo label falls within the top-k of video predictions, we adopt the pseudo label as the final prediction. Otherwise, we utilize the video prediction.
As shown in Table \ref{tab:addlabel}, taking UCF101-C as an example with $\theta$ set to 0.3 and $k$ set to 10, the results show that each label filtering strategy brings further improvement. The best effect is achieved when both strategies are used at the same time.

\begin{table}[t]
\centering
\caption{\dq{Comparison of the impact of different LLMs on AVMIT-C dataset using TANet w.r.t accuracy(\%).}} 
% \vspace{0.2cm}
\resizebox{0.5\textwidth}{!}
{
\begin{tabular}{>{\color{black}}l<{} >{\color{black}}l<{} *{5}{>{\color{black}}c<{}}} % 列格式自动标蓝
\toprule
\multicolumn{1}{>{\color{black}}l<{}}{\multirow{2}{*}{Method}} & \multirow{2}{*}{LLM} & \multicolumn{5}{>{\color{black}}c<{}}{Corruptions} \\ 
\cline{3-7} 
\multicolumn{1}{>{\color{black}}c<{}}{}                        &                      & Gauss & Pepper & Salt  & Shot  & Avg.  \\ 
\hline
Source                                      & -                    & 41.46 & 36.77  & 32.19 &   71.04    & 45.37  \\
ViTTA                                       & -                    & 52.71 & 49.58  & 38.65 & 77.19 & 54.53 \\ 
\hline
\multirow{4}{*}{Ours}                       & GPT-4~\cite{gpt4} (Default)   & 72.40  & 63.85  & 65.73 & 84.06 & 71.51 \\
                                            & Claude-3-sonnet~\cite{claude3} & 73.65 & 64.17  & 61.67 & 84.79 & 71.07 \\
                                            & Llama3~\cite{llama3} & 69.69 & 60.10   & 60.83 & 85.00    & 68.91 \\
                                            & Bing~\cite{bing}     & 71.77 & 58.75  & 57.08 & 84.17 & 67.94 \\ 
\bottomrule
\end{tabular}
}
\label{tab:different_llm}
\end{table}

\begin{figure}[!t]
    \centering 
    \includegraphics[width=\linewidth]{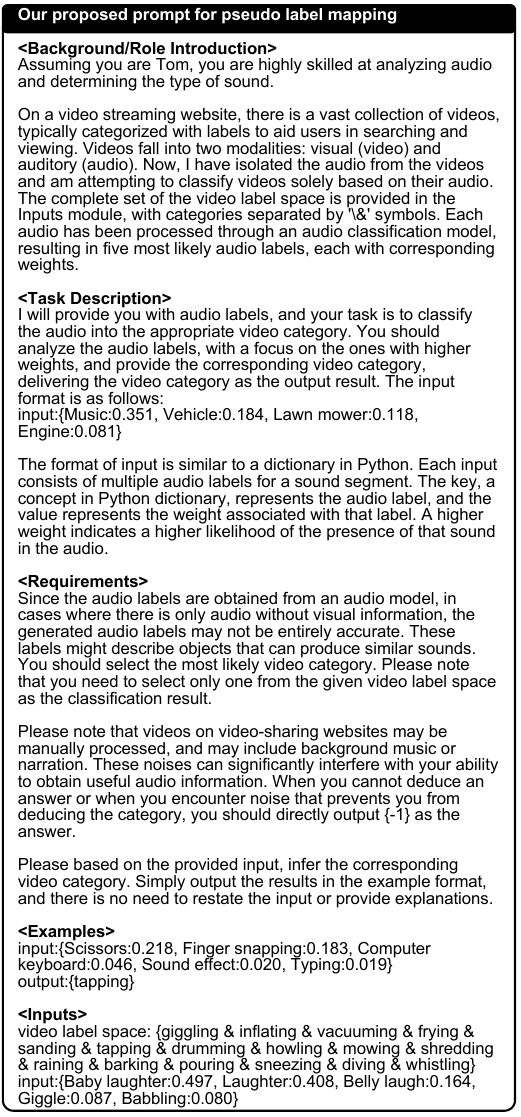}
    % \vspace{-0.6cm}
    \caption{Our proposed prompt for pseudo label mapping.}
    \label{Alg:prompt}
    \vspace{-10pt}
\end{figure}

\subsection{More details about prompt design}
\label{sec:label_mapping}

Our prompt for pseudo label mapping consists of five modules: Background, Task, Examples, Requirements, and Inputs. We furnish contextual information about the task in the Background module, defining the specific task for Large Language Models (LLMs) to execute, and providing multiple illustrative examples. The Requirements module outlines a set of criteria, such as response formatting, to ensure uniformity in the outputs produced by LLMs. In the Inputs module, we supply LLMs with information such as the video label space, the top-k audio labels of the samples, and their corresponding probabilities. This assists LLMs in accurately identifying the most relevant video label based on the content of the sample. The details of our prompt are illustrated in Fig. \ref{Alg:prompt}.

\subsection{\dq{Impact of different LLMs}}
\dq{To assess the impact of different LLMs on the performance of our method, we employ various LLMs for label mapping. The experimental results are presented in Table~\ref{tab:different_llm}. Our findings show that all the tested LLMs effectively support our audio-assisted TTA method. Notably, the performance of the TTA method remained nearly identical when using GPT-4~\cite{gpt4} and Claude-3-Sonnet~\cite{claude3}, suggesting that our approach is not dependent on a specific LLM.}

\subsection{\dq{Impact of different audio models}}
\dq{To evaluate the impact of different audio models on the performance of our method, we further validate our approach by incorporating two other audio backbones: AuM~\cite{AuM} and BEATs~\cite{beats}. As shown in Table~\ref{tab:audio-model}, our audio-assisted TTA method is compatible with these audio backbones, and it significantly improves video TTA performance compared to ViTTA. This demonstrates that our method is not restricted to a specific audio model, highlighting its generalizability.}

\subsection{\dq{Integration with audio-visual models}}
\dq{We combine our method with an audio-visual model, namely, TIM~\cite{tim}, which explicitly models the temporal interaction between audio and visual events in long videos and achieves SOTA performance in multimodal video recognition. Specifically, we train the model using clean training data, then test it under conditions of visual corruption and apply our method for test-time adaptation (TTA). As shown in Table~\ref{tab:audio-visual-model}, our method is compatible with the audio-visual model and improves its TTA performance when facing visual corruption. This combination highlights how our method can enhance the robustness of existing audio-visual models, particularly in challenging scenarios involving visual corruption, further supporting the practical value of our approach.}

\dq{In real-world applications, audio information may not always be available. Our method ensures backward compatibility—if audio is unavailable during testing, the system reverts to the original visual-only performance. On the other hand, audio-visual models would fail in such situations, making our approach more robust in scenarios where audio is intermittently available.
}

\begin{table*}[t]
\centering
\caption{\dq{Performance of different audio models on AVE-C dataset using TANet w.r.t. accuracy (\%).}} 
% \vspace{0.2cm}
\resizebox{\textwidth}{!}
{
\begin{tabular}{>{\color{black}}l<{} *{13}{>{\color{black}}c<{}}}
% \begin{tabular}{lccccccccccccc}
\toprule
\multirow{2}{*}{Model} & \multicolumn{13}{>{\color{black}}c<{}}{Corruptions} \\  \cline{2-14}
% \multirow{2}{*}{Model} & \multicolumn{13}{c}{Corruptions}                                                                                  \\ \cline{2-14} 
                       & Gauss & Pepper & Salt  & Shot  & Contrast & Impulse & Rain  & Zoom  & Motion & Jpeg  & Defocus & H265.abr & Avg.  \\ \hline
Source                 & 35.82 & 35.32  & 23.38 & 60.20 & 20.40    & 35.57   & 56.97 & 42.29 & 68.41  & 48.26 & 50.75   & 51.24    & 44.05 \\
ViTTA                  & 51.00 & 53.23  & 36.07 & 69.65 & 36.07    & 51.54   & 64.43 & 48.01 & 69.65  & 67.66 & 51.24   & 53.73    & 54.36 \\
AST(Ours)              & 62.44 & 63.68  & 51.74 & 74.63 & 47.51    & 59.70   & 70.65 & 62.44 & 70.40  & 70.15 & 61.94   & 58.21    & 62.79 \\
AuM                   & 59.70 & 59.95  & 50.25 & 72.39 & 44.53    & 61.69   & 70.15 & 61.44 & 70.15  & 69.65 & 61.19   & 58.46    & 61.63 \\
BEATs                  & 60.95 & 61.94  & 53.23 & 72.64 & 51.00    & 61.19   & 71.64 & 61.19 & 71.14  & 69.90 & 61.69   & 61.69    & 63.18 \\ \bottomrule
\end{tabular}
}
\label{tab:audio-model}
\end{table*}

\begin{table*}[t]
\centering
\caption{\dq{Performance on the audio-visual TIM model on AVE-C dataset w.r.t. accuracy (\%).}} 
% \vspace{0.2cm}
\resizebox{\textwidth}{!}
{
\begin{tabular}{>{\color{black}}l<{} *{13}{>{\color{black}}c<{}}} % 修改列格式自动标蓝
\toprule
\multicolumn{1}{>{\color{black}}c<{}}{}               & \multicolumn{13}{>{\color{black}}c<{}}{Corruptions} \\ % 多列标蓝
\cline{2-14} 
\multicolumn{1}{>{\color{black}}c<{}}{\multirow{-2}{*}{Method}} 
                       & Gauss          & Pepper         & Salt           & Shot           & Contrast       & Impulse        & Rain           & Zoom           & Motion         & Jpeg           & Defocus        & H265.abr       & Avg.           \\ 
\hline
TIM~\cite{tim}                 & 72.71          & 70.92          & 65.45          & 78.78          & 51.59          & 72.64          & 75.10          & 63.18          & 75.97          & 77.99          & 73.51          & 67.31          & 70.43          \\
TIM+Ours                   & \textbf{73.83} & \textbf{71.07} & \textbf{66.02} & \textbf{78.83} & \textbf{52.49} & \textbf{74.90} & \textbf{77.46} & \textbf{63.53} & \textbf{76.84} & \textbf{78.23} & \textbf{73.56} & \textbf{68.53} & \textbf{71.27} \\ 
\bottomrule
\end{tabular}
}
\label{tab:audio-visual-model}
\end{table*}

\subsection{Training statistics from different datasets}
\label{sec:statistics}

In the study detailed in Table \ref{Tab:statistics}, we examine the efficacy of adapting a model to a test set by employing training statistics derived from another dataset. Our findings demonstrate that adaptation using training statistics from clean datasets consistently enhances model performance. This improvement is observed even when the training statistics are sourced from a dataset different from that of the test set. We observe that utilizing training statistics from either dataset for adaptation to the test set of the other incurs only a minimal performance decrement. This suggests robustness in model performance to variations in training statistics.

\begin{table}[t]
\caption{Mean Top-1 Classification Accuracy (\%) over four types of Noise corruptions, to a test set of AVE-C with train statistics from different datasets. We perform adaptation on TANet~\cite{liu2021tam} and TSM~\cite{Lin_Gan_Han_2020} in combination of different train and test sets across AVMIT and AVE.}
\centering
\scriptsize
  \resizebox{\columnwidth}{!}{
    \begin{tabular}{@{}llcccc@{}}
    % \midrule
        \toprule
        \multirow{2}{*}{Test Set} & \multirow{2}{*}{Backbone} & \multirow{2}{*}{Methods} & \multirow{2}{*}{Source} & \multicolumn{2}{c}{Train Statistics}  \\ \cmidrule{5-6} 
                                  &        &        &        & AVE             & AVMIT          \\ \midrule
        \multirow{4}{*}{AVE-C}    &
        \multirow{2}{*}{TANet}    & ViTTA  & \multirow{2}{*}{38.68}     & 52.49          & 48.63 \\
                                  & & Ours    &                           & \textbf{62.07}            & \textbf{58.09}          \\ \cmidrule{2-6}
        &        \multirow{2}{*}{TSM}      & ViTTA  &  \multirow{2}{*}{34.14}    & 44.16          & 42.73 \\
                                  & & Ours    &                           & \textbf{50.56}            & \textbf{48.20}          \\  
        \bottomrule
    \end{tabular}
    }
    \label{Tab:statistics}
\end{table}

\section{Conclusion}
Existing video TTA methods predominantly rely on visual information, neglecting the potential contribution of audio, which could offer valuable insights for enhancing TTA performance. In this work, we have introduced a novel audio-assisted video TTA framework that incorporates audio signals from videos to improve the generalization ability of video models. Specifically, we have proposed an LLM-based pseudo-label mapping method to address the challenge of aligning audio categories with the video label space, thereby facilitating audio-guided supervision. Additionally, we have introduced an adaptive strategy for determining the number of adaptation steps, which adjusts dynamically based on the data and the specific types of corruptions, optimizing the use of audio-assisted labels. Experimental results across four benchmark datasets have validated the effectiveness of our approach. Our findings provide a promising solution for integrating audio information in TTA contexts, enhancing the robustness of video models under diverse conditions.

% \section*{Acknowledgments}
% This work was partially supported by National Natural Science Foundation of China (NSFC) under Grants 62202311, the Guangdong Basic and Applied Basic Research Foundation under Grants 2023A1515011512, Excellent Science and Technology Creative Talent Training Program of Shenzhen Municipality under Grant RCBS20221008093224017, Key Scientific Research Project of the Department of Education of Guangdong
% Province 2024ZDZX3012.
% \ys{This should be a simple paragraph before the References to thank those individuals and institutions who have supported your work on this article.}

% \appendix
% \section*{Our proposed prompt for LLM summarization}

\bibliographystyle{IEEEtran}
\bibliography{main}

\vfill
\end{document}